\documentclass[journal]{IEEEtran}

\usepackage[T1]{fontenc}
\usepackage[utf8]{inputenc}
\usepackage{graphicx}
\usepackage{amsmath,amssymb}
\usepackage{booktabs}
\usepackage{xcolor}
\usepackage[hidelinks,hypertexnames=false]{hyperref}
\hypersetup{
  pdftitle={Rapid Debris-Volume Estimation from
            Post-Hurricane Aerial Imagery},
  pdfauthor={Kooshan Amini, Jamie E. Padgett, Guha Balakrishnan},
  pdfsubject={Hurricane debris volume estimation from post-event aerial
              imagery},
  pdfkeywords={remote sensing, hurricane debris, debris volume, monocular
               depth estimation, foundation models}
}
\usepackage{cite}
\usepackage{overpic}
\usepackage{threeparttable}
\usepackage{float}

\graphicspath{{figures/main/}{figures/si/}}

\begin{document}
\lefthyphenmin=3 \righthyphenmin=3
\setlength{\emergencystretch}{6pt}

\title{Rapid Debris-Volume Estimation from
Post-Hurricane Aerial Imagery}

\author{Kooshan~Amini\textsuperscript{1},
        Jamie~E.~Padgett\textsuperscript{1,3},
        and~Guha~Balakrishnan\textsuperscript{2,3}
\thanks{\textsuperscript{1}Department of Civil and Environmental
Engineering, Rice University, Houston, TX, USA.}%
\thanks{\textsuperscript{2}Department of Electrical and Computer
Engineering, Rice University, Houston, TX, USA.}%
\thanks{\textsuperscript{3}Ken Kennedy Institute, Rice University, Houston,
TX, USA.}%
\thanks{Corresponding author: Jamie E. Padgett (e-mail:
jamie.padgett@rice.edu).}%
}

\maketitle

\begin{abstract}
Hurricane debris removal is planned, contracted, and federally reimbursed on
the basis of volume estimates, yet operational practice still relies on
parametric forecasts with 41--90\% documented over-estimation or on
truck-load tallies that arrive only after hauling begins. We present
DebrisHeightNet, a segmentation-conditioned monocular debris-height network
that estimates spatially explicit debris volume from a single pass of
post-event aerial RGB imagery, the kind of survey routinely flown within
days of a hurricane landfall. We train only a lightweight
1.08\,M-parameter head on top of two frozen vision foundation models. This
head regresses height from a Depth Anything~V2 backbone, conditioned on the debris
segmentation of CLIPSeg-debris from our prior work. Because no
post-hurricane debris-height ground truth exists, we synthesize the training
target by confidence-weighted LiDAR--monocular fusion (CW-LMF), designed to
suppress non-debris LiDAR returns. This fused target is a constructed
supervision signal rather than ground truth, so we corroborate it against
external references rather than claiming it as truth. A region-level power-law calibration, driven by each region's low-density debris fraction, converts model volume into an
estimate of the reported hauled debris with quantified uncertainty. Across ten regions
spanning five hurricanes and three states, the uncalibrated model agrees
with an independent uncrewed-aerial-vehicle (UAV) survey of the training region at
Spearman $\rho = 0.87$ and lands within 30\% of the reported record
where the Hazus and FEMA-hybrid parametric forecasts over-predict it by
2.7--4.8$\times$. Deployment requires no LiDAR, no ground access,
and no second flight, so the method can produce spatially explicit volume
estimates wherever single-pass post-event imagery is flown.

\end{abstract}


\section{Introduction}\label{sec:intro}

\IEEEPARstart{D}{ebris} generated by hurricanes is among the largest and most
time-critical burdens of disaster recovery in the United States. A single
landfall can leave behind millions of cubic meters of mixed construction,
demolition, and vegetative debris, and its removal routinely dominates early
recovery budgets and timelines~\cite{fema_debris_guide}. Debris quantities are
also drive administrative decisions. Under the federal Public Assistance program,
debris removal is a distinct reimbursement category whose mission scale
and contracting are scoped on the basis of \emph{estimated
quantities}~\cite{fema_papg}, beginning with the preliminary damage
assessments that precede and inform a declaration request, which must be
submitted within 30 days of the incident~\cite{cfr206_36}. Yet, the estimates
that drive these decisions remain coarse. Operational practice relies either
on parametric forecasting models that map storm intensity and building
inventory to regional debris tonnage, or on post-hoc truck-load tallies that
arrive only after hauling is underway. A recent review of disaster-waste
quantification reports an error of roughly 30\% for the widely
used U.S. Army Corps of Engineers (USACE) method and a documented debris
over-estimation of 41--90\% for the FEMA hurricane
model~\cite{marchesini2021}. Load-ticket accounting avoids these modeling errors but  accrues only as hauling proceeds and is a documented source of billing fraud and associated federal oversight findings~\cite{gao2020,gao2026disaster}. An
independent, spatially explicit, and rapidly available estimate of debris
volume would strengthen every link of this chain, from mission scoping
to reimbursement audit.

\begin{figure*}[t]
\centering
\includegraphics[width=\textwidth]{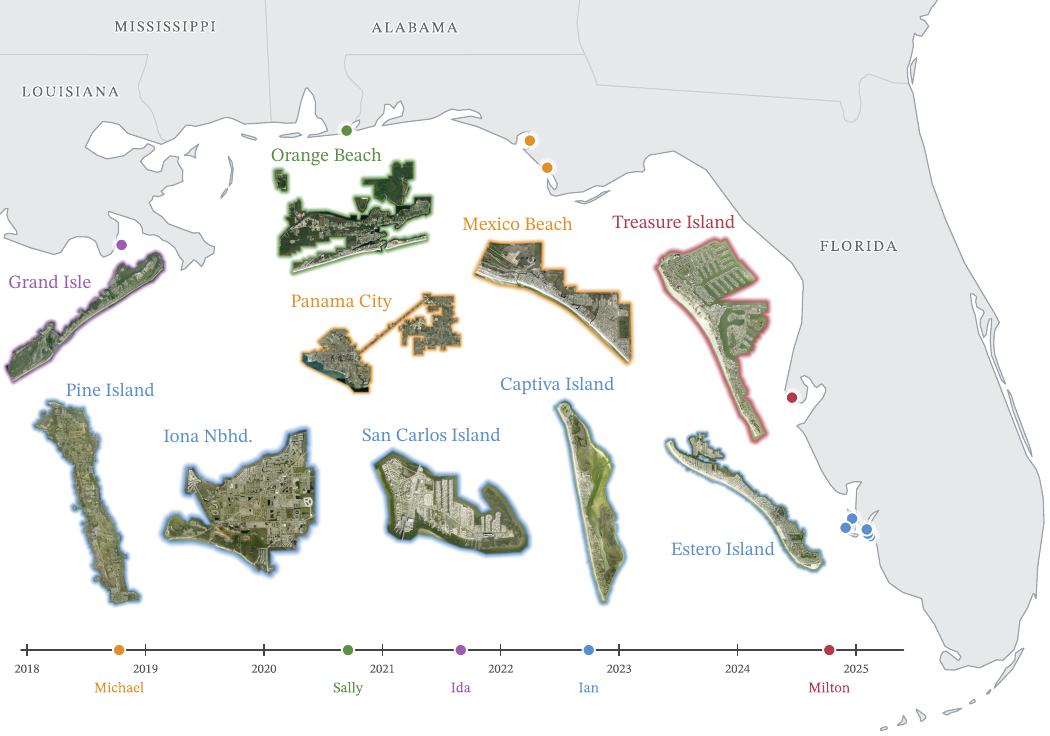}
\caption{\textbf{Study regions.} Ten hurricane-affected coastal regions spanning five
hurricanes (2018--2024) and three U.S. states: Hurricane Ian (Pine Island,
Estero Island, San Carlos Island, Captiva Island, and Iona, all FL), Michael
(Panama City and Mexico Beach, FL), Ida (Grand Isle, LA), Sally (Orange
Beach, AL), and
Milton (Treasure Island, FL). Region cutouts are ringed in their storm's
color, which also marks each region's location on the Gulf-coast locator and
on the event timeline. Estero Island is the single training region; no other
region contributes training labels. Basemap imagery: Esri World Imagery; boundaries:
Natural Earth.}
\label{fig:study_area}
\end{figure*}

Remote sensing should be able to provide such an estimate, but existing
approaches each fall short of operational needs. Parametric and
data-driven forecasting models, from the Hazus hurricane model and FEMA's
field guidance to recent statistical variants, require no imagery
at all, yet they produce only county- or region-level totals, without finer
spatial detail, and carry the accuracy limitations noted
above~\cite{fema_debris_guide,hazus_hurricane,gonzalezduenas2023}. Three-dimensional measurement methods use airborne LiDAR, synthetic aperture radar (SAR), or multi-view photogrammetry to reconstruct the post-event surface or to difference it against a pre-event one~\cite{koyama2016,jiang2022,cheng2024}. They measure volume directly, but they depend on sensors or repeat flight geometries that are rarely tasked at wide-area scale within days of a landfall. Deep-learning methods for post-disaster mapping, by contrast, operate on the
post-event aerial and satellite RGB imagery that is routinely available
after storms. Building-damage benchmarks such as xBD, and the damage- and
debris-mapping networks and datasets that followed
them, segment and classify what this imagery shows~\cite{gupta2019xbd,cheng2021,braik2024,kaur2023,rahnemoonfar2023}, but their outputs are two-dimensional extent and class, not volume. The National Oceanic and Atmospheric Administration's Emergency Response Imagery (NOAA ERI) exemplifies both the
opportunity and the obstacle: it is flown within days of major U.S. hurricane landfalls (Fig.~\ref{fig:study_area}) and images the affected coastal areas at decimeter resolution~\cite{noaa_eri}, but it is a single-pass, monocular RGB product, with no stereo pairs and no elevation channel, so the third dimension needed for volume has, to our knowledge, never been extracted from it.

This paper closes that gap by advancing debris mapping from detection alone
to
quantification. In prior work, we fine-tuned the vision--language
segmentation model CLIPSeg~\cite{luddecke2022clipseg} into CLIPSeg-debris, which delineates low- and high-density hurricane debris in post-event aerial imagery~\cite{amini2025}. Here we present \emph{DebrisHeightNet} (Fig.~\ref{fig:pipeline}), a segmentation-conditioned monocular debris-height network that estimates per-50\,m-grid hurricane debris volume from a single post-event aerial RGB image, trained on labels generated by confidence-weighted LiDAR--monocular fusion (CW-LMF). We reuse frozen vision foundation models rather than train task-specific
perception from scratch~\cite{radford2021clip,kirillov2023sam}. A frozen
Depth Anything V2 backbone~\cite{yang2024depthv2} supplies dense but
relative monocular depth, the frozen CLIPSeg-debris segmentation supplies
conditioning, and we train only a lightweight 1.08\,M-parameter head to
estimate debris height in meters, a data-efficiency choice made for
out-of-distribution transfer.

Two further obstacles shape the method. First, no post-hurricane debris-height ground truth exists. Post-event LiDAR, when available, arrives days to months later, is gappy, and is contaminated by returns from standing vegetation and intact roofs. We therefore synthesize the training target with CW-LMF, which fuses post-event LiDAR with calibrated monocular depth under a per-pixel agreement confidence designed to suppress spurious non-debris returns. This target is itself a model output. Without LiDAR truth at the imagery epoch it is corroborated rather than proven, so we treat its corroboration as a deliberate component of the study design: we check it against an independent drone survey, reported hauled quantities, and a class-by-height decomposition against the source LiDAR. Second, image-derived volume and hauled volume differ systematically across debris regimes, because hauled records include material that a single-pass image cannot register. The model's own output is
therefore left uncalibrated. As usage guidance, we provide a region-level power-law calibration that converts accumulated model volume into an estimate of the total hauled quantity, reported with an explicit out-of-sample uncertainty (a 90\% prediction interval spanning a factor of about 3.6 in either direction); the calibration's predictor is insensitive to how much debris-free area the analyst's region of interest (ROI) includes. At deployment the pipeline has the advantage of consuming imagery alone.

We evaluate the method across ten hurricane-affected regions spanning five
storms (Ian, Michael, Ida, Sally, Milton) and three U.S. states
(Fig.~\ref{fig:study_area}), against reported hauled-debris records assembled
from public-records requests, FEMA project data, and municipal sources, and
against an
independent UAV structure-from-motion survey. The quantitative results,
including a head-to-head comparison against the parametric models used in
practice, are presented in Sec.~\ref{sec:results}.

The contributions of this work are as follows.
\begin{itemize}
  \item \textbf{A monocular debris-height model.} DebrisHeightNet regresses
  per-pixel debris height from a single post-event aerial RGB image by
  conditioning a lightweight trained head on frozen
  foundation-model depth and debris segmentation; to our knowledge it is the
  first
  single-image method for hurricane debris height and volume, where prior
  debris
  work is two-dimensional or requires active sensing.
  \item \textbf{CW-LMF supervision without clean ground truth.} A
  confidence-weighted, LiDAR-primary fusion that synthesizes a debris-isolated
  height target where no post-hurricane debris-height truth exists.
  \item \textbf{Volume calibration with quantified uncertainty.} A
  region-level
  power-law calibration that maps model
  volume
  to reported hauled volume with an explicit 90\%
  prediction interval, compared against the parametric models
  used in practice.
  \item \textbf{Broad multi-region evaluation.} The method is evaluated on ten
  regions across five hurricanes and three states, with an independent UAV
  cross-check; to our knowledge this is the broadest multi-region evaluation
  of
  debris-volume estimation reported to date.
\end{itemize}
Section~\ref{sec:related} reviews related work;
Section~\ref{sec:method} details the methodology;
Section~\ref{sec:data} describes the study regions and data; and
Sections~\ref{sec:results}--\ref{sec:conclusion} present the results,
discussion, and conclusions.

\section{Related Work}\label{sec:related}

\subsection{Disaster-Debris Quantification}\label{sec:related_debris}
Existing debris-quantity estimates come from three families of methods.
\emph{Predictive models} forecast debris before or independently of
observation: FEMA's Hazus hurricane model derives building and tree debris
from wind fields and building inventories~\cite{hazus_hurricane}, agency field
guides codify per-structure formulas and field estimation
practice~\cite{fema_debris_guide,fema329}, recent data-driven variants
learn debris quantities from storm intensity, exposure, and past-event
records~\cite{gonzalezduenas2023}, and uncertainty-aware workflows infer
building-debris volume from AI-classified damage states and
inventory~\cite{cheng2024workflow}. These produce actionable regional totals
but no spatial detail, and their documented accuracy limits were noted
above~\cite{marchesini2021}. \emph{Operational accounting} quantifies debris
from the removal process itself, using load tickets and disposal records,
as in
a post-Harvey reconciliation of collected flood
debris~\cite{bekkaye2023}; such records are the accepted reference but only
materialize weeks to months into hauling, and a 2026 federal audit of the
Helene response found that affected localities received debris-progress
information months late~\cite{gao2026disaster}. \emph{Remote-sensing measurement}
extracts the deposited volume from data. Examples include polarimetric
stereo-SAR height
differencing after the 2011 Tohoku tsunami~\cite{koyama2016}, drone
photogrammetry with deep learning and GIS for construction-and-demolition
debris stockpiles~\cite{jiang2022}, AI-assisted aerial-photogrammetry
frameworks for disaster debris~\cite{cheng2024}, and post-hurricane
assessments pairing photogrammetric differencing against pre-event LiDAR
with learned debris detection~\cite{aggarwal2025helene}. A head-to-head
comparison of satellite, aerial, UAV, and terrestrial-laser options for
debris quantification confirms that accuracy and coverage trade off across
these sensing choices~\cite{bekkaye2024application}. These methods measure
volume
directly but presuppose active sensors, dedicated multi-view flights, or
site-scale campaigns. None delivers spatially explicit debris volume from the
single-pass RGB imagery that is actually typically flown over the affected coast
within days of
a U.S. landfall, and validation against hauled quantities across multiple
events remains rare.

\subsection{Deep Learning and Foundation Models for Post-Disaster
Mapping}\label{sec:related_dl}
The xBD dataset~\cite{gupta2019xbd} standardized building-damage assessment
from satellite imagery and seeded a family of increasingly capable
damage-mapping networks, from post-hurricane aerial
classification~\cite{cheng2021} and multi-view fusion~\cite{khajwal2023} to
GIS-integrated large-scale mapping~\cite{braik2024} and hierarchical
transformers~\cite{kaur2023}; UAV-scale benchmarks such as
RescueNet~\cite{rahnemoonfar2023} and CRASAR-U-DROIDs~\cite{manzini2024crasar}
extend the task to higher-resolution post-storm scenes. More recently, vision
foundation models such as CLIP~\cite{radford2021clip} and Segment Anything
(SAM)~\cite{kirillov2023sam} have entered
Earth-observation and civil-engineering workflows~\cite{rs_foundation_survey}:
prompt-based adaptations of SAM for remote-sensing instance
segmentation~\cite{chen2024rsprompter} and for post-disaster damage
evaluation~\cite{zhao2025vipde}, vision--language pipelines that extract a
physical measurement from imagery (building lowest-floor
elevation)~\cite{ho2025}, rapid regional post-hazard assessment of structures
and transportation infrastructure~\cite{yang2025rapid}, and self-supervised
monocular depth estimation for construction scenes~\cite{shen2023}. At the
same time, a recent multi-hazard vision--language benchmark finds that both
generic and remote-sensing--specific foundation models still struggle on
disaster-focused quantitative tasks~\cite{wang2025disasterm3}. For
debris specifically, our prior work fine-tuned
CLIPSeg~\cite{luddecke2022clipseg} into CLIPSeg-debris, which segments low-
and high-density debris in post-hurricane aerial imagery and attains a Dice
score of 0.86 on a hurricane excluded from training~\cite{amini2025}. Across
this literature, however, the output is a class map, an extent, or a
point measurement. Post-disaster deep learning, including debris
mapping, stops at two dimensions, and the volumetric quantity that debris
missions are actually planned around has remained out of reach.

\subsection{Monocular Depth and Height Estimation from Overhead
Imagery}\label{sec:related_height}
Monocular depth estimation has matured into a foundation-model stack:
cross-dataset training delivered robust zero-shot relative
depth~\cite{ranftl2022midas}, dense vision transformers improved its
resolution~\cite{ranftl2021dpt}, and the Depth Anything series scaled it to
tens of millions of
images~\cite{yang2024depthv1,yang2024depthv2,depthanything_v3};
Metric3D~v2 targets zero-shot metric depth
given camera intrinsics~\cite{hu2024metric3d}. Because such models output
relative (affine-invariant) depth, using them for measurement requires a
scale anchor. Recent aerial benchmarking further shows that even
metric-depth variants degrade at overhead viewpoints without
adaptation~\cite{song2026aerialmetric}. In parallel, a remote-sensing literature regresses height
directly from single overhead images. Examples include
IM2HEIGHT~\cite{mou2018im2height},
IM2ELEVATION~\cite{liu2020im2elevation},
IMG2nDSM~\cite{karatsiolis2021img2ndsm}, and HTC-DC
Net~\cite{chen2023htcdc}, with transferable representations, transformer
variants, synthetic pre-training, and weak supervision from imperfect height
labels extending
generalization~\cite{xiong2023benchmark,chen2024heightformer,zhao2023contrastive,song2024synrs3d,chen2026enhancing}.
These methods target buildings and terrain, and they lean on LiDAR-derived
normalized digital surface models (nDSMs) as supervision, a dependence that
becomes the central obstacle when no suitable LiDAR exists. Closest to our
setting, depth-foundation backbones have been adapted to general single-view
remote-sensing height estimation~\cite{hong2025depth2elevation} and to
canopy-height estimation~\cite{cambrin2024canopy}, and global canopy mapping
supervises dense height regression with spaceborne LiDAR
footprints~\cite{lang2023canopy}; sparse LiDAR returns have likewise been
used to correct monocular height predictions~\cite{song2025lidarguided}.
Debris, however, differs from buildings and canopy in the properties that
drive height estimation: it is transient, geometrically irregular, and
co-located with the
standing vegetation and intact structures that contaminate post-event LiDAR.
To our knowledge, no single-image method estimates deposited-debris height or
volume (single-image photoclinometry has recovered debris-flow erosion
volume from terrain relief~\cite{zhang2025debrisflow}, a different quantity),
none couples height regression with debris segmentation to output the
mission-relevant quantity, and none confronts the supervision problem that
no
debris-height ground truth exists at the imagery epoch. These are the three
gaps this paper addresses with
DebrisHeightNet, its debris-conditioned volume integration, and the CW-LMF
fused training target.

\section{Methodology}\label{sec:method}

\subsection{Problem Formulation and Overview}\label{sec:method_overview}
\begin{figure*}[t]
\centering
\includegraphics[width=\textwidth]{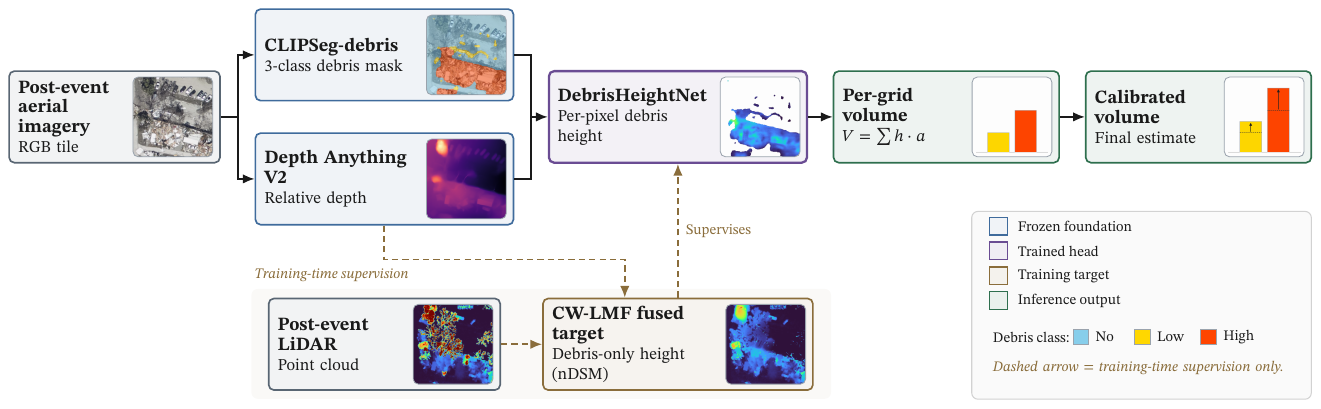}
\caption{\textbf{Pipeline overview with a worked example cell.} Two frozen foundation
models read the post-event RGB tile: CLIPSeg-debris produces the three-class
debris mask and DA-V2 produces dense relative depth. The trained
DebrisHeightNet head converts these into per-pixel metric debris height,
which is integrated over the cell (per-grid volume) and calibrated to
reported hauled volume at
the region level (calibrated volume). The dashed path is
outside the deployment flow. The fused nDSM from confidence-weighted
LiDAR--monocular fusion (CW-LMF, Sec.~\ref{sec:method_cwlmf})
supervises the height head during training (and separately builds
the evaluation references of Sec.~\ref{sec:data_protocol}), so no LiDAR is
required at deployment.}
\label{fig:pipeline}
\end{figure*}

We estimate hurricane debris volume on a regular 50\,m grid. Each grid cell
is observed by a single post-event aerial RGB tile, and the target output
is the debris volume within the cell's 2{,}500\,m\textsuperscript{2}
footprint. Fig.~\ref{fig:pipeline} traces the pipeline from this tile
onward. The tile enters two frozen foundation models in parallel:
CLIPSeg-debris segments it into no-, low-, and high-density debris, and
Depth Anything~V2 (DA-V2) produces dense but relative depth.
DebrisHeightNet (DHN), the only trained component, merges the two streams
and regresses per-pixel metric debris height. Integrating these heights
over a cell's debris-classified
pixels gives the uncalibrated per-grid volume. A region-level calibration,
fitted
to historical reported hauled-debris records, then converts accumulated
model
volume into the final calibrated estimate of the total volume to be hauled,
including
material not visible from the air, so that the organizations responsible
for debris missions obtain totals consistent with the quantities they
historically record, with quantified uncertainty. The dashed branch of
Fig.~\ref{fig:pipeline} exists only at training time. There, CW-LMF fuses
post-event LiDAR with the calibrated monocular depth into the debris-only
height target that supervises the head (Sec.~\ref{sec:method_cwlmf}).

This division of labor is the core design choice. The frozen models carry
the generic perception burden, and we train only a lightweight
1.08\,M-parameter head on the $\approx$3{,}600-cell labeled region, a
data-efficiency choice made for out-of-distribution transfer; the
development comparisons of Sec.~\ref{sec:results_ablations} are consistent
with it. Deployment consumes imagery alone. Beyond the training target,
LiDAR serves only the fused evaluation references of
Sec.~\ref{sec:data_protocol}.

\subsection{DebrisHeightNet: Segmentation-Conditioned Height
Regression}\label{sec:method_dhn}
We reuse the segmentation front-end from our prior work without
modification. CLIPSeg-debris labels every pixel as no-debris, low-density
debris, or high-density debris~\cite{amini2025}, and enters the height head
as a one-hot conditioning tensor. Conditioning on the mask lets the head
apply different height priors to dispersed thin debris and to piled rubble,
and confines learning capacity to the debris classes that matter for volume.

\begin{figure*}[t]
\centering
\includegraphics[width=\textwidth]{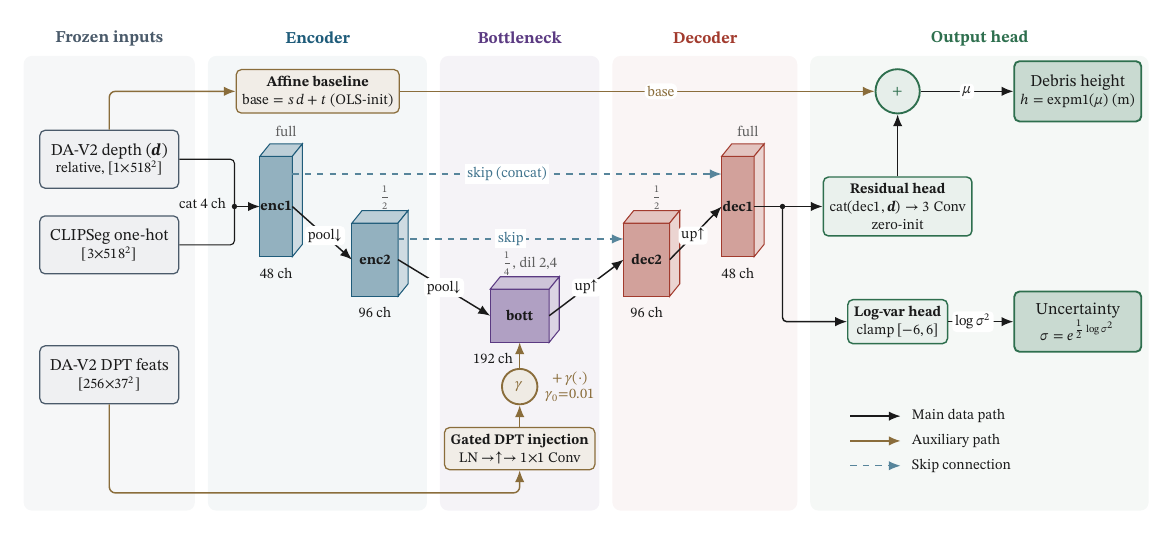}
\caption{\textbf{DebrisHeightNet architecture.} Frozen inputs (DA-V2 relative depth,
CLIPSeg-debris one-hot mask, pooled DPT decoder features) feed an
OLS-initialized affine depth baseline and a compact U-Net encoder--decoder
(48/96/192 channels, dilated bottleneck) with gated feature injection
(gate initialized at 0.01). A zero-initialized residual head corrects the
baseline and a parallel head outputs per-pixel log-variance, yielding a
heteroscedastic height prediction ($\mu$, $\log\sigma^2$) in
$\log(1+h)$ space. Only the head's 1{,}084{,}885 parameters are trained.}
\label{fig:architecture}
\end{figure*}

DebrisHeightNet (Fig.~\ref{fig:architecture}) receives three frozen
inputs per tile: the DA-V2 ViT-L relative depth map
($1\times518\times518$)~\cite{yang2024depthv2}, the CLIPSeg-debris one-hot
mask ($3\times518\times518$), and intermediate decoder features of the frozen
depth model ($256\times37\times37$, average-pooled from the dense
prediction transformer (DPT) decoder~\cite{ranftl2021dpt}). A learnable affine
baseline $s\!\cdot\!D+t$, initialized by ordinary least squares (OLS) against the
training target, anchors the relative depth $D$ to an approximate metric
scale; a compact four-channel U-Net encoder--decoder (48/96/192 channels,
dilated bottleneck) refines it, with the pooled decoder features injected at
the bottleneck through a learned gate initialized near zero (0.01) so that
training opens the pathway only if it helps. A zero-initialized residual head
adds the learned correction to the affine baseline, and a parallel head
predicts a per-pixel log-variance (clamped to $[-6,6]$), making the output
heteroscedastic: a height estimate $\mu$ and an uncertainty for every pixel.
(The variance output serves as a training-time noise model,
Sec.~\ref{sec:method_loss}; its calibration as a deployable per-pixel
uncertainty is not evaluated in this work.)
We model heights in $\log(1+h)$ space and map them back by $\exp(\cdot)-1$
at inference, which linearizes the strongly right-skewed debris-height
distribution. In total 1{,}084{,}885 parameters are trained; the backbone
remains frozen. DA-V2 was retained over its newer successor
\cite{depthanything_v3} because it produced structurally more coherent depth
on post-event aerial scenes in our backbone comparison
(Sec.~\ref{sec:results_ablations}).

\subsection{Debris-Focused Loss}\label{sec:method_loss}
Debris heights are right-skewed (many small values with a long tail of tall
piles), spatially sparse, and operationally asymmetric, because
under-predicting a large pile is costlier than over-predicting scattered
material. The training loss therefore combines six standard components from heteroscedastic regression and monocular depth estimation, each
addressing one of these properties. A
heteroscedastic Gaussian negative log-likelihood on
$(\mu,\log\sigma^2)$, the standard objective for
regression under input-dependent noise~\cite{kendall2017uncertainties}, fits
the noisy fused target while learning a per-pixel
uncertainty, and an asymmetric mean-absolute-error (MAE) term
penalizes under-prediction $2.5\times$ more than over-prediction, reflecting
the operational cost asymmetry. Per-pixel class and height-proportional
weights counter the
extreme imbalance between debris and background pixels. A
gradient-matching term, standard in monocular depth
estimation~\cite{ranftl2022midas}, preserves the sharp pile boundaries that
pixelwise losses blur, and a differentiable per-cell volume term aligns the
training objective with
the quantity the application consumes. A small regularizer on the predicted
log-variance prevents variance collapse. We fixed all weights once
against the training region's validation split during development and held
them constant thereafter; none was tuned on any evaluation region. We
quantify the
contribution of the principal choices in the ablations of
Sec.~\ref{sec:results_ablations} and give the complete equations with every
numeric weight in the Supplementary Information.

\subsection{CW-LMF: Synthesizing the Training Target}\label{sec:method_cwlmf}
No post-hurricane height ground truth exists for hurricane debris. Post-event
LiDAR, when available, is metrically accurate but gappy (dropouts over
water, dark roofs, and thin debris) and contaminated by
returns from standing vegetation and intact structures that are not debris;
monocular depth is dense and anchored to the visible surface but only
relative. Confidence-weighted LiDAR--monocular fusion (CW-LMF) combines the
two into a debris-isolated training target (Supplementary
Fig.~\ref{fig:si_cwlmf}). Three principles govern the fusion: LiDAR is primary wherever it is present and plausible; per-pixel trust is decided by the measurable \emph{agreement} between the two sources rather than by either source alone; and where LiDAR is absent or implausible, the calibrated monocular surface substitutes.

Let $L$ denote the raw LiDAR nDSM of a cell (post-event LiDAR surface minus
the pre-event bare-earth digital elevation model (DEM), water masked, clamped non-negative) and $D$ the DA-V2 relative depth. After integer registration (grid search up to
4\,px maximizing gradient correlation, yielding a registration-quality map $Q$), the depth is calibrated to metric scale by an iteratively reweighted robust linear fit $D_{\mathrm{cal}} = aD + b$ whose weights combine $Q$ with a per-pixel agreement confidence $C$. The confidence is a fixed-weight combination of five
complementary LiDAR--depth agreement measures (structural similarity, gradient
correlation, height agreement, edge overlap, and a stable-extreme
bonus). A conflict score then quantifies residual disagreement,
\begin{equation}
S \;=\; w_h\,\bar{R} \;+\; w_g\,\bar{G} \;+\; w_q\,(1-Q) \;+\;
w_c\,(1-C),
\label{eq:conflict}
\end{equation}
where $\bar{R}$ and $\bar{G}$ are the normalized height and gradient
mismatches of $L$ versus $D_{\mathrm{cal}}$ and the fixed weights
$w_{(\cdot)}$ emphasize the height mismatch, the primary symptom of a
spurious non-debris return; cells of $S$ above an
adaptive robust threshold form a conflict mask. The per-pixel blend weight
(the share of the monocular estimate) is
\begin{equation}
\alpha \;=\;
\begin{cases}
0.25\, C\, Q\, (1-S) & \text{LiDAR present and agreeing,}\\[2pt]
1 & \text{LiDAR gap or hard conflict,}
\end{cases}
\label{eq:alpha}
\end{equation}
smoothed and feathered at boundaries, and the fused target is
\begin{equation}
F \;=\; (1-\alpha)\,L \;+\; \alpha\,D_{\mathrm{cal}}.
\label{eq:fuse}
\end{equation}
The design is LiDAR-first. In agreeing interior overlap the monocular share
is capped
at 25\%, although boundary feathering can locally exceed the cap. Where the
two sources strongly conflict, the LiDAR
return is treated as spurious non-debris structure (canopy, intact roof)
and
replaced by the calibrated depth. This conflict rule is a design assumption
rather than a per-pixel arbitration, and it is designed to strip tall
non-debris returns from the target. (In the
deployed Estero target the delivered LiDAR rasters contained no voids, so
the gap branch of Eq.~(\ref{eq:alpha}) never activated; the target's changes
stem from blending and conflict replacement.) We fixed the full
configuration (the agreement and conflict weights, thresholds, and blending
parameters, listed in the Supplementary Information) from these design
principles and fusion diagnostics on the training region, before any
evaluation-region comparison, and never tuned it against any evaluation
quantity or reported record. We claim the design logic, not parameter
optimality. The fused target remains a model output: we do not claim it is
more accurate than raw LiDAR, and without LiDAR truth at the imagery epoch
this remains unproven. We therefore corroborate it against an independent
drone survey, reported hauled volumes, and a class-by-height decomposition
against the source LiDAR (Secs.~\ref{sec:results_corroboration} and~\ref{sec:discussion}).

\subsection{Training, Inference, and Volume
Integration}\label{sec:method_training}
We train on a single region, Estero Island, Florida (Hurricane
Ian, 2022), which offers the strongest combination of co-located data in
our set: complete
fused-LiDAR coverage and the most extensively reconciled
reported-debris record. We split the 3{,}645 cells of the Estero grid block
that carry a valid fused target into 2{,}767 training, 525
validation, and 353
held-out test cells ($\approx$75/15/10\%) by whole $5\times5$ blocks of
consecutive cell identifiers. Roughly a thousand of these cells contain no
mapped debris, the background context
the head must also learn. Because the identifier sequence
follows the acquisition numbering rather than the island's exact geometry,
this blocking reduces but does not eliminate adjacency between splits, so
the held-out error may be somewhat optimistic relative to a fully spatially
isolated split.\footnote{The training grid was delineated
independently of the evaluation region of interest and extends slightly
beyond it, which is why its 3{,}645 cells differ from the 3{,}178-cell
extent of Table~\ref{tab:master}.}
Table~\ref{tab:training_config} summarizes the optimization settings;
training runs
on a single NVIDIA RTX~6000 Ada GPU. We select the best epoch by a
composite criterion combining validation height MAE with
class-level volume agreement, which yields epoch 1138
(validation MAE 0.760\,m;
held-out test MAE 0.582\,m).

\begin{table}[!t]
\centering
\caption{Training configuration of DebrisHeightNet.}
\label{tab:training_config}
\footnotesize
\setlength{\tabcolsep}{5pt}
\renewcommand{\arraystretch}{1.12}
\begin{threeparttable}
\begin{tabular}{@{}l l@{}}
\toprule
\textbf{Setting} & \textbf{Value} \\
\midrule
\multicolumn{2}{@{}l}{\emph{Optimization}} \\
Optimizer              & AdamW (weight decay 0.005) \\
Learning-rate schedule & OneCycleLR, max LR $7\times10^{-5}$ \\
Batch size             & 48 \\
Epochs                 & 2000 \\
Mixed precision        & enabled (AMP) \\
Gradient clipping      & max-norm 1.0 \\
Weight averaging       & EMA, decay 0.9995 \\
\midrule
\multicolumn{2}{@{}l}{\emph{Data and split}} \\
Input tile size        & $518$\,px (upsampled 256\,px cell tile) \\
Split protocol         & identifier blocks ($5\times5$), seed 42 \\
Validation / test fraction & 0.15 / 0.10 \\
Train / validation / test cells & 2{,}767 / 525 / 353 (Estero Island) \\
\midrule
\multicolumn{2}{@{}l}{\emph{Model selection}} \\
Best-epoch criterion   & composite (val.\ MAE $+$ volume terms) \\
Selected epoch         & 1138 \\
Validation MAE         & 0.760\,m \\
Held-out test MAE      & 0.582\,m \\
Trainable parameters   & 1{,}084{,}885 \\
\bottomrule
\end{tabular}
\begin{tablenotes}[flushleft]\scriptsize
\item AMP: automatic mixed precision; EMA: exponential moving average; MAE:
mean absolute error.
\end{tablenotes}
\end{threeparttable}
\end{table}

At inference, the 256\,px cell tile is upsampled to the backbones' native
input resolution, DA-V2 is run through its full-precision interface at
$518\times518$, the trained head produces $\mu$, and heights are mapped to
meters and re-rasterized to the 256\,px cell grid. The uncalibrated cell
volume
sums heights over the debris-classified pixels of a cell $g$,
\begin{equation}
V_g \;=\; a_{\mathrm{px}} \!\!\sum_{p\,\in\, g,\; s_p \in
\{\text{low},\,\text{high}\}} \!\!\max(\hat{h}_p, 0),
\label{eq:volume}
\end{equation}
where $\hat{h}_p = \exp(\mu_p) - 1$ is the predicted height, $s_p$ is the
CLIPSeg-debris class of pixel $p$, and
$a_{\mathrm{px}} = (50/256)^2 \approx 0.038$\,m\textsuperscript{2} is the
pixel footprint. Throughout the paper, ``uncalibrated'' volumes are these
direct sums with no post-hoc adjustment; the only downstream correction is
the region-level calibration of Sec.~\ref{sec:method_calibration}, and each
reported volume is labeled as either uncalibrated or calibrated.

\subsection{Region-Level Calibration to Reported
Debris}\label{sec:method_calibration}
Image-derived volume and hauled volume are related but not identical
quantities, because hauled records include material invisible from the air
(interior content considered waste, debris deposited later) and vary in scope, while image-derived volume
depends on what the sensor resolves in each debris regime. We therefore close
the loop with a region-level calibration fitted against reported
hauled-debris records (Sec.~\ref{sec:data_reference}). The organizations
that scope, contract, and reimburse debris
missions plan around total hauled quantities, so the calibration
converts what the imagery measures into that operational quantity, absorbing
what a single-pass image cannot capture, rather than
altering the model itself. Across regions, a single predictor $m$ explains
the ratio of reported to model volume well. This predictor is the mean
low-density debris fraction over a region's debris-bearing cells, and it is
insensitive, in the precise sense given below, to how the
analyst's region of interest (ROI) is
drawn. Tall, piled debris is detected well wherever it occurs.
Under-detection is driven by thin, dispersed material, and $m$ measures
how much of that faint low-density signature the imagery registers.
Regions with small $m$ are therefore the regimes a single-pass optical method
under-detects most, and the calibration multiplier
\begin{equation}
\hat{c}(m) \;=\; 0.0238\, m^{-1.152}
\label{eq:powerlaw}
\end{equation}
maps accumulated model volume to reported volume, with $m$ expressed as a
dimensionless fraction (the figures display it in percent),
$\hat{V}^{\mathrm{cal}}_r = \hat{c}(m_r)\sum_{g \in r} V_g$. Because $m$ is
averaged over debris-bearing cells only, it is insensitive to how much
debris-free area an analyst's ROI includes: enlarging an ROI into debris-free
surroundings leaves it unchanged, and only the inclusion of further
debris-bearing cells can move it. We chose the predictor by out-of-sample
performance:
among the candidate region descriptors compared under leave-one-out (LOO)
validation, $m$ attains the best held-out $R^2$, area-normalized fractions
are ROI-dependent and weaker, land-cover covariates underperform,
and non-logarithmic functional forms overfit (Supplementary
Fig.~\ref{fig:si_calibration}). We report the
calibration by its out-of-sample behavior, a 90\% prediction interval
spanning a factor of about 3.6 in either direction for a new region, with
the full fit statistics
in
Sec.~\ref{sec:results_calibration}. Regions whose $m$ falls below roughly
1\% are flagged as a low-reliability tier; in this faint-signature regime,
observed multipliers exceed
$5\times$.

\section{Study Regions and Data}\label{sec:data}

\subsection{Regions and Imagery}\label{sec:data_regions}
The study covers ten hurricane-affected coastal regions across five
landfalling storms and three U.S. states (Fig.~\ref{fig:study_area}):
Hurricane Ian, 2022 (Pine Island, Estero Island, San Carlos Island, Captiva
Island, and the Iona neighborhood of Fort Myers, all Florida); Hurricane
Michael, 2018 (Panama City and Mexico Beach, Florida); Hurricane Ida, 2021
(Grand Isle, Louisiana); Hurricane Sally, 2020 (Orange Beach, Alabama); and
Hurricane Milton, 2024 (Treasure Island, Florida). The set intentionally
mixes surge-dominated barrier islands, wind-dominated panhandle towns, and
low-destruction dispersed-debris regimes. Estero Island is the single
training and calibration-anchor region; every other region contributes no
training labels and is used for evaluation, although all ten regions
enter the region-level calibration of Sec.~\ref{sec:method_calibration}.

Each region is tessellated into non-overlapping 50\,m grid cells, and
each cell is rasterized to a $256\times256$ tile ($\approx$0.2\,m ground
sample distance) from NOAA Emergency Response Imagery, the single-pass RGB
product flown after major U.S. hurricane landfalls. For every
region we use the earliest available post-event flight,
one to two days after landfall in all ten regions. For the training
target, and at evaluation time for the fused references of
Sec.~\ref{sec:data_protocol}, we additionally use post-event airborne LiDAR
(USACE and NOAA/NGS
coastal-mapping campaigns distributed through NOAA Digital Coast, NAVD88)
and 1\,m pre-event bare-earth DEMs (USGS 3DEP and
NOAA products). This asymmetry is deliberate. LiDAR serves training and
evaluation
only, and deployment consumes imagery alone. Post-event LiDAR acquisition
lagged landfall by 11--24 days (Ian) and 14--25 days (Michael), and where a
campaign exists it covers between 0\% and 100\% of a region's debris-bearing
cells; Grand Isle's post-Ida campaign was flown more than 13 months after
landfall, during ongoing cleanup, so height references built from it
under-capture the
event-time debris. Complete per-region
provenance (flight identifiers, LiDAR campaigns and lags, DEM sources,
coordinate systems, and realized coverage) is given in Supplementary
Table~\ref{tab:provenance}. Fig.~\ref{fig:region_examples} shows one example
cell per storm
across the data layers. The examples make the supervision problem concrete,
as the raw LiDAR nDSM carries tall returns from intact roofs and standing
vegetation adjacent to the debris, which the fused target suppresses in
these examples while
retaining the piles themselves. The fusion also visibly regularizes the raw
surface where the two sources agree: it cleans building edges, suppresses
isolated noisy returns, and softens view-angle and registration
artifacts. The resulting target is therefore both debris-isolated and
smoother
than either input alone.

\begin{figure*}[t]
\centering
\includegraphics[width=\textwidth]{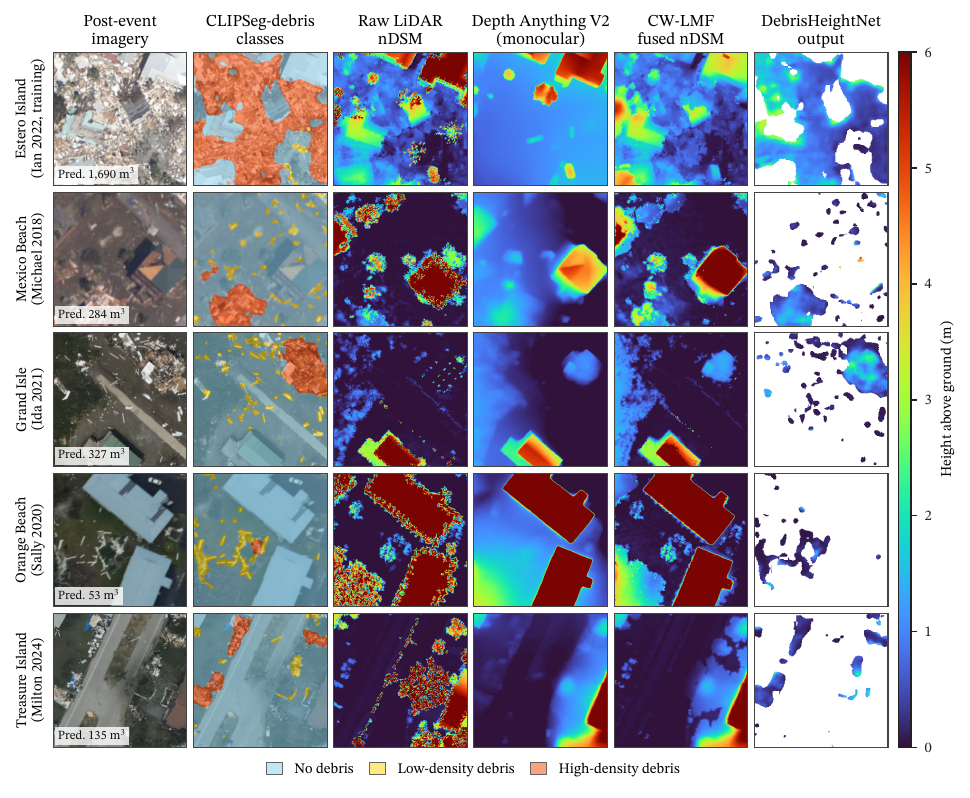}
\caption{\textbf{Example cells, one region per storm (rows: Estero Island, training;
Mexico Beach; Grand Isle; Orange Beach; Treasure Island).} Columns: post-event
NOAA imagery; CLIPSeg-debris three-class mask; raw post-event LiDAR nDSM;
DA-V2 relative depth (affine-aligned for display); CW-LMF fused surface;
DebrisHeightNet height prediction. Heights share a common 0--6\,m
scale. Regions differ in LiDAR coverage and acquisition lag
(Sec.~\ref{sec:data_regions}); the LiDAR and fused columns are shown for
context, and only Estero's fused targets are used for training.}
\label{fig:region_examples}
\end{figure*}

\subsection{Reference and Validation Datasets}\label{sec:data_reference}
The external reference for calibration and
region-level evaluation is the volume of debris actually collected, assembled
per region from various sources. These include per-truck load records released by Lee County under a
Florida public-records request, spatially filtered
to each of the five Ian
regions; FEMA Public Assistance figures (Panama City, Orange Beach); a FEMA
project-data construction-and-demolition line item (Mexico Beach); and
contractor or municipal reporting for Grand Isle and for Treasure Island,
whose record
is a combined Helene--Milton contract. Because the Treasure Island imagery
(flown 11 October 2024) also postdates Hurricane Helene's 26 September 2024
landfall, the imagery and the record capture the same two-storm deposition. For each region, the value we denote
\emph{Reported} is the median of the reconciled region-scope estimates (one
to four sources per region); the per-source catalogue, scope of each record,
vegetation share where published, and minimum--maximum spans appear in
Supplementary Table~\ref{tab:real_debris_sources}. These records are
imperfect and heterogeneous by nature, mixing
vegetation-inclusive and construction-and-demolition-only scopes and
municipal versus ROI extents. We therefore carry that heterogeneity
explicitly as
per-region spans rather than treating Reported as exact truth.

One additional dataset supports the
training-target corroboration without entering training or calibration:
an NSF Natural Hazards Engineering Research Infrastructure (NHERI)
RAPID uncrewed-aircraft survey of Estero Island, flown roughly
three weeks after Ian's landfall (DesignSafe project
PRJ-4211~\cite{designsafe_prj4211}), which provides a
structure-from-motion (SfM) surface from a different sensor at a different
epoch; we compare against it at grid level in
Sec.~\ref{sec:results_corroboration}. It is a
top-surface product that retains standing vegetation and intact structures
and does not match the imagery epoch; it corroborates rather than
proves the training target (Sec.~\ref{sec:method_cwlmf}).

\subsection{Evaluation Protocol and Metrics}\label{sec:data_protocol}
Model performance is evaluated at three spatial levels, each establishing a
different property of the estimates. \emph{Aggregate volume ratios} measure
scale agreement for the totals that debris
missions consume. Each ratio divides total model volume by one of three
references: the Reported record, a region's fused reference, or the drone
survey. Because a near-unity ratio can arise from offsetting
errors, ratios are always read together with their composition. For these
comparisons, the fusion of Sec.~\ref{sec:method_cwlmf} is additionally
applied to each evaluation region whose LiDAR coverage supports it, yielding
a per-region fused reference used only as an internal consistency check
(training uses Estero alone; Sec.~\ref{sec:method_training}).
\emph{Grid-level association} (Spearman rank correlation and Pearson
correlation of log volumes) measures whether the model orders cells
correctly and is insensitive to overall scale. This ranking property matters for prioritizing debris-management
zones, pickup deployments, and routing, hence particularly relevant for our context. \emph{Pixel-level spatial
correlation} probes the finest-grain arrangement of predicted height within
cells; it is the weakest of the three signals, as the method is designed for
integrated cell volume rather than per-pixel height placement. Every
reported number is labeled in-distribution (Estero Island) or
out-of-distribution (any other region). Computationally, the trained head
adds 1.08\,M parameters
on top of the frozen backbones; training converged in about 24 hours on a
single NVIDIA RTX 6000 Ada GPU, and inference runs on the same single GPU.
Measured end to end on cached imagery, segmentation processes about 16
cells per second, while depth and feature extraction (about 7 cells per
second) and the height head (over 40 cells per second) run only on the
debris-flagged subset (Table~\ref{tab:master}). The full San Carlos Island
region (697 cells) completed in a measured 2.1 minutes, and the same stage
rates put the largest region (Panama City, 45{,}394 cells, 13{,}354
debris-bearing) at roughly 1.5 hours.

\section{Results}\label{sec:results}
\setcounter{dbltopnumber}{2}

\subsection{Cross-Region Debris Volume}\label{sec:results_crossregion}
The first question for a model trained in a single region is whether its
volume estimates hold up across storms, coastal built environments, and debris regimes it has
never seen. Table~\ref{tab:master} addresses this directly, comparing
uncalibrated DHN volume with the reported hauled-debris record in all ten
regions; the discussion below additionally draws on each region's
comparison against its fused LiDAR reference, for the regions with enough
LiDAR coverage to build one. Its \emph{Reported} column is the reconciled median
that the calibration of Sec.~\ref{sec:method_calibration} is fit to, and
\emph{Rep./DHN} is the corresponding reported-to-model multiplier: near 1,
the uncalibrated model already matches the record; far above 1, it
under-detects.

\begin{table*}[t]
\centering
\caption{Ten-region comparison of uncalibrated DebrisHeightNet (DHN) volume
with per-region reported hauled-debris records. Reported: median of the
region's reconciled estimates (sources in Supplementary Table~\ref{tab:real_debris_sources}). Rep./DHN:
reported-to-model multiplier. $N$ grids: full-ROI\,/\,debris-bearing cells.
Volumes in m\textsuperscript{3}.}
\label{tab:master}
\footnotesize
\setlength{\tabcolsep}{7pt}
\renewcommand{\arraystretch}{1.1}
\begin{tabular}{@{}l r r r r r@{}}
\toprule
\textbf{Region} & \textbf{Area (km\textsuperscript{2})} &
\textbf{$N$ grids (ROI\,/\,debris)} & \textbf{DHN (m\textsuperscript{3})} &
\textbf{Reported (median, m\textsuperscript{3})} & \textbf{Rep./DHN} \\
\midrule
\multicolumn{6}{@{}l}{\emph{Hurricane Ian -- Lee County, FL (2022)}} \\
Pine Island        & 80.80  & 32{,}320\,/\,5{,}603  & 227{,}931 & 442{,}576 & 1.94 \\
Estero Island$^{\ast}$ & 7.95 & 3{,}178\,/\,2{,}154 & 239{,}343 & 187{,}868 & 0.78 \\
San Carlos Is.     & 1.74   & 697\,/\,427           & 60{,}919  & 100{,}122 & 1.64 \\
Captiva Island     & 8.52   & 3{,}406\,/\,299       & 5{,}951   & 52{,}186  & 8.77 \\
Iona Nbhd.         & 39.55  & 15{,}821\,/\,4{,}941  & 218{,}090 & 342{,}667 & 1.57 \\
\midrule
\multicolumn{6}{@{}l}{\emph{Hurricane Michael -- FL Panhandle (2018)}} \\
Panama City        & 113.49 & 45{,}394\,/\,13{,}354 & 408{,}188 & 857{,}705 & 2.10 \\
Mexico Beach       & 7.65   & 3{,}059\,/\,1{,}654   & 112{,}274 & 373{,}134 & 3.32 \\
\midrule
\multicolumn{6}{@{}l}{\emph{Hurricane Ida -- Jefferson Parish, LA (2021)}} \\
Grand Isle         & 27.09  & 10{,}836\,/\,1{,}661  & 62{,}632  & 276{,}587 & 4.42 \\
\midrule
\multicolumn{6}{@{}l}{\emph{Hurricane Sally -- Baldwin County, AL (2020)}} \\
Orange Beach       & 44.30  & 17{,}720\,/\,1{,}405  & 6{,}609   & 243{,}283 & 36.81 \\
\midrule
\multicolumn{6}{@{}l}{\emph{Hurricane Milton -- Pinellas County, FL (2024)}} \\
Treasure Island    & 6.48   & 2{,}590\,/\,662       & 18{,}226  & 146{,}795 & 8.05 \\
\bottomrule
\multicolumn{6}{@{}l}{\rule{0pt}{2.2ex}\scriptsize $^{\ast}$Trained only on
Estero Island cells (Sec.~III-E); no other region contributes training
labels.}
\end{tabular}
\end{table*}

\begin{figure*}[t]
\centering
\includegraphics[width=\textwidth]{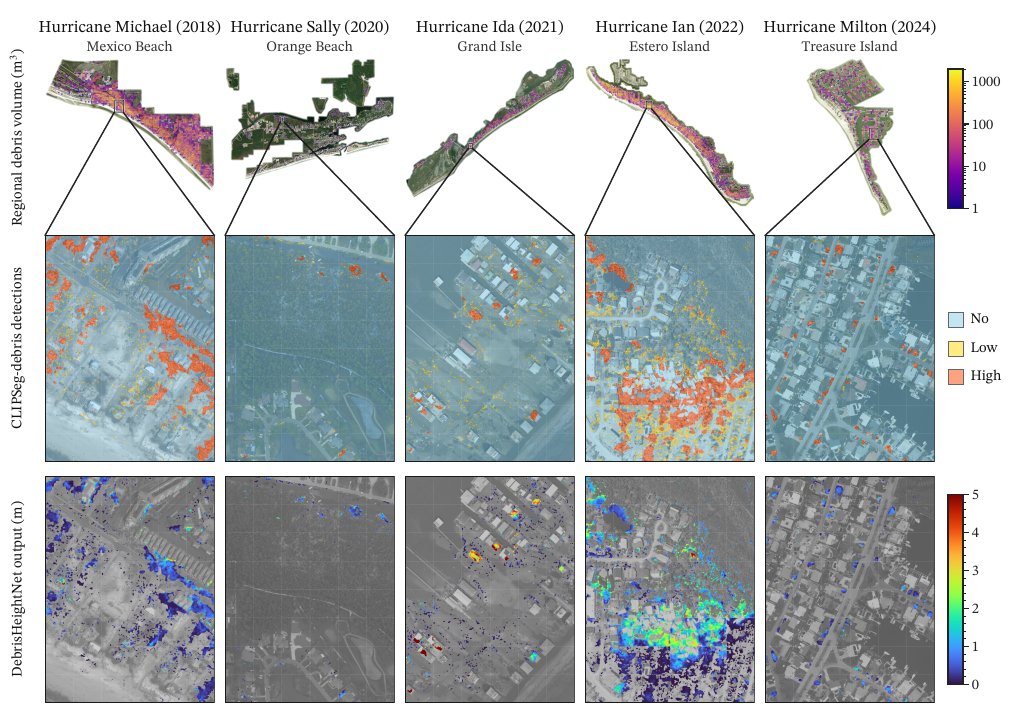}
\caption{\textbf{Spatial predictions, one example region per storm (left to right:
Mexico Beach, Orange Beach, Grand Isle, Estero Island, Treasure Island).}
Top: whole-ROI per-cell volume map (logarithmic scale,
1--2{,}000\,m\textsuperscript{3}) over post-event imagery. Middle and bottom:
a zoomed high-debris pocket with CLIPSeg-debris detections and
DebrisHeightNet heights (0--5\,m). The Orange Beach column is the
dispersed-debris failure regime discussed in
Sec.~\ref{sec:results_crossregion}.}
\label{fig:spatial}
\end{figure*}

\begin{figure*}[t]
\centering
\includegraphics[width=\textwidth]{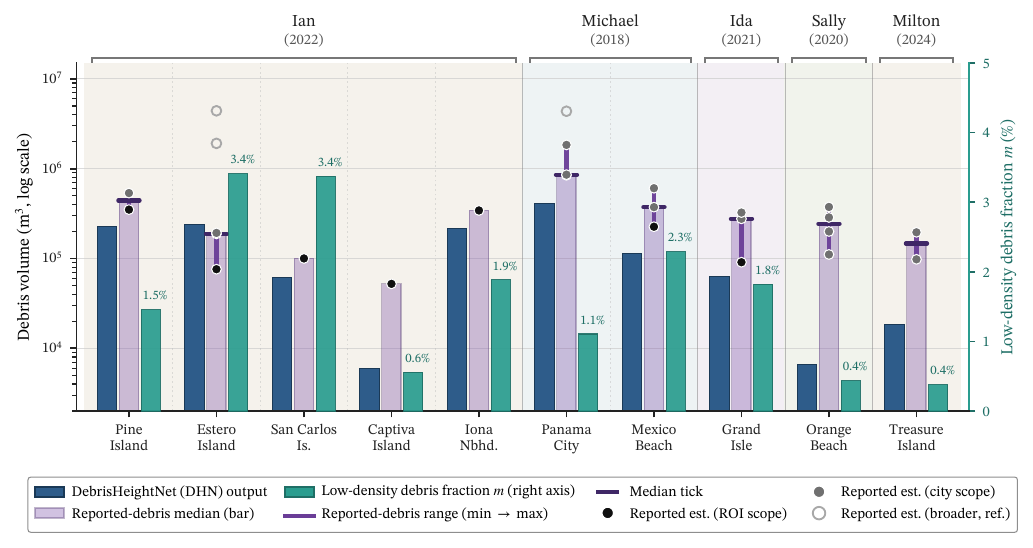}
\caption{\textbf{Per-region estimates.} Uncalibrated DHN volume (bars) against the
reported hauled-debris median, span, and individual sources (markers), with
the low-density debris fraction $m$ (right axis) that drives
the calibration. Spans cover the reconciled region-scope sources; the full
catalogue, including broader-scope records, is in Supplementary
Table~\ref{tab:real_debris_sources}. Regions are grouped by hurricane; note
the logarithmic volume axis.}
\label{fig:cross_region}
\end{figure*}

We read the table in two steps. First, we check \emph{internal consistency}.
In three of the
five regions with near-complete LiDAR coverage
($\gtrsim$90\%; per-region coverage in Supplementary Table~\ref{tab:provenance}), the model
reproduces the
reference within 5--14\%: model-to-reference ratios are 1.05 and 1.14 at the
out-of-distribution San Carlos and Captiva Islands and 1.06 in-distribution
at Estero
Island. Mexico Beach under-shoots its reference (0.65), a deviation
we cannot attribute to a documented cause. Grand Isle
overshoots (1.53) in the direction its cleanup-lagged reference implies. The
LiDAR, flown more than 13 months late, under-captures event-time debris, so
an overshoot
of some size is expected there. Second, we check \emph{agreement with the
reported
record}. The Rep./DHN multipliers stratify cleanly by debris regime rather
than scattering randomly. At Estero the uncalibrated model is already within
30\% of the record (0.78); the surge-affected Ian regions and Panama City
cluster at 1.57--2.10; Mexico
Beach and Grand Isle sit at 3.3--4.4; and the low-destruction,
dispersed-debris regimes collapse to 8--37$\times$ (Captiva, Treasure
Island, Orange Beach). This is the regime dependence that the low-density
predictor
$m$ captures and the calibration of Sec.~\ref{sec:method_calibration}
corrects. Fig.~\ref{fig:cross_region} shows this stratification graphically,
plotting each region's estimate against the reported median and the span of
its underlying sources, with the predictor $m$ overlaid on a second axis.
The inverse relationship is visible directly. The regions whose estimates
fall farthest below their records are precisely those with the smallest
low-density fractions. This previews the calibration that
Sec.~\ref{sec:results_calibration} quantifies.
Finally, Fig.~\ref{fig:spatial} maps the predictions themselves as whole-ROI
per-cell volume maps with zoomed detection and height detail, one region per
storm. Its Orange Beach panel doubles as the failure
case of genuinely sparse, thin debris that single-pass optical sensing
under-detects.

\subsection{Calibration to Reported Hauled Debris}\label{sec:results_calibration}
\begin{figure*}[t]
\centering
\includegraphics[width=\textwidth]{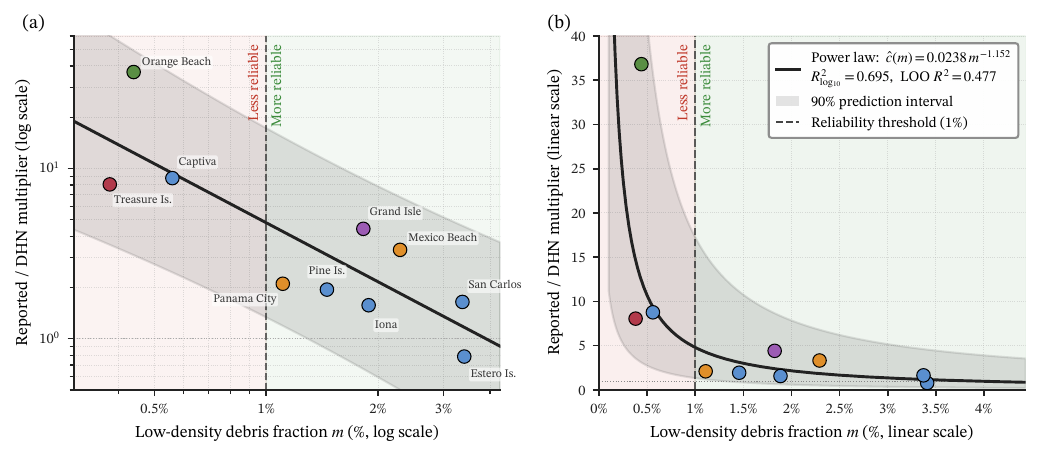}
\caption{\textbf{Region-level calibration.} Reported/DHN multiplier versus the
low-density debris fraction $m$ (log--log and linear views),
with the fitted power law $\hat{c}(m) = 0.0238\,m^{-1.152}$ and its
90\% prediction interval ($m$ as a fraction in the fit; displayed in
percent). Circle colors encode each region's hurricane, using the same
color code as Fig.~\ref{fig:study_area}. The vertical line at $m = 1\%$
marks the
low-reliability tier: regions left of it carry observed multipliers above
5$\times$.}
\label{fig:calibration}
\end{figure*}

The calibration of Sec.~\ref{sec:method_calibration} is useful only if it
generalizes to regions outside its fit, so we report it by its out-of-sample
behavior. Fig.~\ref{fig:calibration} shows the fitted power law of
Eq.~(\ref{eq:powerlaw}), which links the per-region reported-to-model
multiplier to the low-density debris fraction $m$; both of its views show
the same trend: the fainter a region's low-density signature, the
larger the multiplier the record implies. Under leave-one-out
validation the power law explains about half of
the between-region variance ($R^2 = 0.477$). The 90\% prediction interval
for a new region is multiplicative: at the predictor mean it runs from the
estimate divided by 3.6 to the estimate multiplied by 3.6, and it widens to
a factor of about 4.1 at the edges of the observed support
(Student-$t$, $n = 10$; in-sample
$R^2_{\log_{10}} = 0.695$, nominal $p < 0.01$; the shaded band of
Fig.~\ref{fig:calibration}). The predictor
and functional form were selected on the same ten regions, so the
leave-one-out statistic and the nominal significance are themselves
selection-optimistic. The interval also
reflects regression scatter only; the heterogeneity of the underlying
records (Sec.~\ref{sec:data_reference}) is folded into that scatter rather
than modeled separately. Under the same leave-one-out protocol, applying the
predicted multiplier reduces the median multiplicative error of the regional
totals from 2.7$\times$ (uncalibrated) to 1.9$\times$ and the $\log_{10}$
RMSE from 0.72 to 0.33, while the median absolute percentage error is
essentially unchanged ($\approx$61\% versus 62\%). The calibration thus corrects
the regime-scale bias rather than fine percentage accuracy, and its largest
held-out errors fall in the low-reliability tier (Orange Beach,
4.8$\times$). In the faint-signature regime ($m$ below roughly
1\%), observed multipliers reach 8--37$\times$ (Captiva 8.8$\times$,
Treasure Island 8.0$\times$, Orange Beach 36.8$\times$) against fitted
values of 9--15$\times$; estimates there are flagged as a low-reliability
tier rather than given unqualified corrections. For context, the
parametric
practice it would complement reports an error of $\sim$30\% under ideal
conditions
and 41--90\% documented over-estimation for hurricanes~\cite{marchesini2021};
the present calibration carries a wider band on far less input data, while
adding the spatial detail and the explicit, quantified reliability tier that
parametric forecasts cannot provide.

\subsection{Corroboration Against External
References}\label{sec:results_corroboration}
Because the fused training target is corroborated rather than proven
(Sec.~\ref{sec:method_cwlmf}), the estimates must also be checked against
references external to the target's construction; two such checks
are available at the training region.

\begin{figure*}[t]
\centering
\begin{overpic}[width=0.46\textwidth,percent]{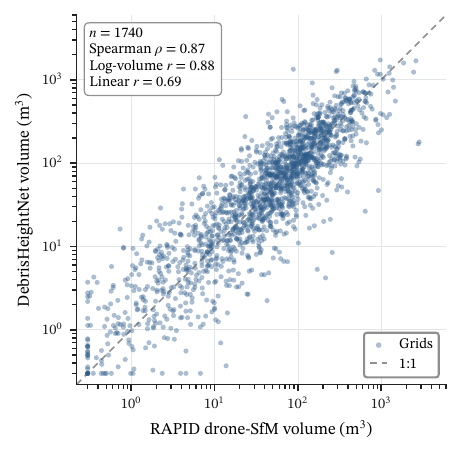}
  \put(3,96.5){\small (a)}
\end{overpic}%
\hfil
\begin{overpic}[width=0.50\textwidth,percent]{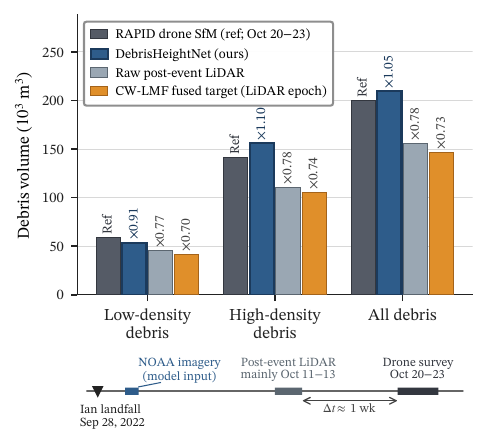}
  \put(2.5,88.8){\small (b)}
\end{overpic}
\caption{\textbf{UAV cross-check at Estero Island.} (a)~Per-cell DHN versus drone-SfM
volume on reference-valid cells ($n = 1{,}740$; log--log). (b)~Aggregate
volume on the common footprint for the drone reference, DHN, raw post-event
LiDAR, and the CW-LMF fused target, by debris class, with the acquisition
timeline (landfall Sep~28; NOAA imagery Sep~30; LiDAR Oct~11--20, 99\% of
returns on Oct~11--13; drone
survey Oct~20--23).}
\label{fig:drone}
\end{figure*}

The first is the independent drone survey of
Sec.~\ref{sec:data_reference}. Against it, the model attains Spearman
$\rho = 0.87$ on the $n = 1{,}740$ cells with a valid drone reference
(Fig.~\ref{fig:drone}(a); log-volume Pearson $r = 0.88$; linear $r = 0.69$,
comparable to the
raw-LiDAR reference's $\approx$0.70 at the same task), and the aggregate
model-to-drone ratio over the full common footprint is 1.05 (1.10 for
high-density,
0.91 for low-density debris; Fig.~\ref{fig:drone}(b)). The near-unity
aggregate partly reflects
offsetting per-height-band differences, because the drone surface retains standing
vegetation and structures that the model excludes. We therefore lead with
the rank
agreement, which is robust to such scale offsets. (Reference validity is
decided by a rule adopted after inspection of the drone surfaces but
computable without model outputs: cells whose mean or
99th-percentile drone height over mapped debris falls below 0.10\,m are
excluded as failed reconstructions, 212 of the 1{,}952
common-footprint cells; the correlations use the remaining 1{,}740 cells,
the aggregate
ratios keep the full footprint, where near-zero cells contribute little
volume, and the exclusion moves Spearman $\rho$ by less than 0.01.) At the finest grain,
per-pixel spatial
correlation between predicted and drone heights over mapped debris is modest
(median $\approx$0.34), consistent with the scoping of
Sec.~\ref{sec:data_protocol}. The reliable output is integrated cell volume,
not per-pixel pile placement.

\begin{figure}[t]
\centering
\includegraphics[width=\columnwidth]{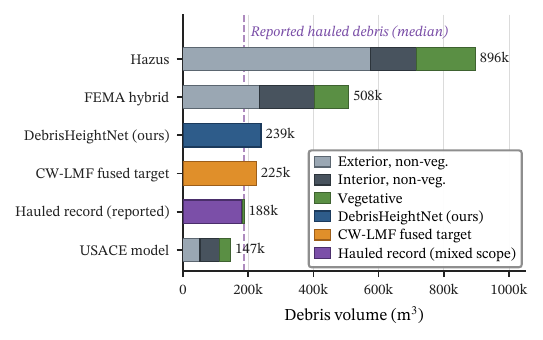}
\caption{\textbf{Method comparison at Estero Island:} incumbent parametric forecasts
(with their exterior/interior/vegetative composition, where the method
reports it) versus the CW-LMF fused target and the uncalibrated
DebrisHeightNet output (ours), against the reported hauled record (dashed:
the reconciled median used throughout the paper). Objective volumes only.}
\label{fig:method_comparison}
\end{figure}

The second check places the estimates next to the parametric forecasts used
in practice, on the one region where every method and a reconciled record
coexist (Fig.~\ref{fig:method_comparison}). Unlike the drone survey, the
reported record here is also one of the ten calibration responses
(Sec.~\ref{sec:method_calibration}), so this check is external to the
training target but not independent of the calibration. Implementations of the Hazus
hurricane debris methodology
\cite{hazus_hurricane}, a FEMA hybrid variant, and the USACE per-structure
method~\cite{fema329}, each driven by per-building damage assessments from
an integrated pre-/post-storm dataset for the same
area~\cite{designsafe_prj5700}, forecast approximately 896{,}000,
508{,}000, and
147{,}000\,m\textsuperscript{3} respectively, against the reported record of
187{,}868\,m\textsuperscript{3} (the reconciled median used in
Table~\ref{tab:master}). The first two over-predict by
factors of 4.8 and 2.7, respectively, while the USACE figure is dominated by
an interior-loss
component that the haul records do not resolve. The CW-LMF fused
target (225{,}019\,m\textsuperscript{3}) and the uncalibrated DHN output
(239{,}343\,m\textsuperscript{3}) land 20\% and 27\% above the record,
respectively.
We show objective volumes only; the comparison is not a ranking exercise,
but it shows that an
imagery-measured estimate lands near the reported record where the Hazus
and FEMA-hybrid
forecasts deviate severalfold. Together with the
decomposition discussed in Sec.~\ref{sec:discussion}, these checks
form the
corroboration base of the fused training target; none is event-epoch
truth, and we treat them as corroborating, not confirmatory.

\subsection{Ablations and Design Choices}\label{sec:results_ablations}
The design choices of Sec.~\ref{sec:method} were adopted for measured
reasons, and this subsection quantifies the principal ones. The comparisons
form a development ladder rather than controlled re-runs (single seed,
shared test set), so we read them as effect sizes, not hypothesis tests.
Three choices
carry most of the performance: replacing water-masked
absolute targets with the water-masked relative formulation produced
the largest single improvement of the development ladder (held-out test MAE
to 0.672\,m); log-space training with gradient matching brought it to
0.569\,m; and the volume-aware epoch criterion accepted a small height-MAE
increase (0.582\,m) in exchange for better class-level aggregate volume
agreement, the
quantity the application consumes. In the development comparisons, which are
single-run like the rest of the ladder, higher-capacity variants that
improved
in-distribution fit degraded out-of-distribution transfer; the
lightweight trained head was therefore retained. For the backbone, DA-V2 ViT-L was
compared
against three variants of its newer successor on our imagery
($n = 3{,}854$ tiles of the 4{,}016-tile Estero grid block; this
population exceeds the 3{,}645 training cells because the comparison needs
no fused target). DA-V2 produced the most structurally coherent depth
(0.304 versus 0.295 for the best Depth Anything 3 variant), the lowest
depth-inversion rate, and the highest reliable fraction, and was retained
(Supplementary Fig.~\ref{fig:si_backbone}). The full calibration-predictor and
functional-form comparisons, including land-cover covariates, support the
choices of Sec.~\ref{sec:method_calibration} (Supplementary Fig.~\ref{fig:si_calibration}).

\section{Discussion}\label{sec:discussion}
Three questions frame the discussion: why a monocular model can measure
debris at all, what the capability offers operationally, and where its
limits lie.

Decomposing the fused training target against the raw LiDAR it modifies
(3{,}490 cells of the Estero evaluation extent; Supplementary
Fig.~\ref{fig:si_redistribution}) explains
why segmentation-conditioned monocular height estimation can measure
debris. Fusion raises the short bands
where debris piles live. Debris-class volume in the 0.5--2\,m bands
increases by
22--65\% as blending with calibrated monocular depth lifts under-measured
short returns. Fusion also suppresses
the tall bands dominated by standing vegetation and intact structure,
removing 81--93\% of debris-class volume in the 5--15\,m bands. The net
effect retains 65--91\% of
class volume depending on class. This is the signature expected of a
height-selective filter separating debris from co-located non-debris
relief, and it supports the interpretation that a monocular model trained on
the
fused target measures debris rather than landscape. We emphasize the
scope of this claim. The decomposition describes a mechanism consistent with
the design intent; it is not proof that the fused surface is more accurate
than raw LiDAR (Sec.~\ref{sec:method_cwlmf}).

Operationally, everything the pipeline needs at
deployment, a single pass of post-event RGB, already exists within days
of a major U.S. hurricane landfall, and processing a region takes minutes
to about 1.5 hours on a single GPU (Sec.~\ref{sec:data_protocol}), well
inside the 30-day window allowed
for major-disaster declaration requests~\cite{cfr206_36}. The output is not a regional total but a per-50\,m
map, which supports zone-level mission planning, progress tracking against
load tickets, and independent cross-checks of reimbursement claims, the
oversight gap documented for debris
operations~\cite{gao2020,gao2026disaster}. Against the
accuracy limits of the parametric status quo documented in
Sec.~\ref{sec:intro}, the method offers an
image-conditioned estimate rather than an inventory-based forecast. Furthermore, it provides an
explicit uncertainty band and a
reliability tier that tells an emergency manager when not to trust
it.

The method has five main limitations in its current form. First,
\emph{cross-sensor scale drift}: the model's height scale is learned from a
single imagery source, and imagery with different acquisition
characteristics (sensor, altitude, radiometry, processing) can shift the
recovered scale. A uniform scale shift leaves the ranking of cells
unchanged, and refitting the height scale on a small sample from the new
sensor is the natural remedy; larger changes in imagery characteristics
remain untested. Systematic evaluation on further
sources, including satellite imagery, remains future work, although the
drone cross-check already shows encouraging agreement against a reference
from a
different sensing technology and epoch
(Sec.~\ref{sec:results_corroboration}).
Second, the training target is corroborated, not proven
(Sec.~\ref{sec:method_cwlmf}); the two surface-based corroboration legs are
vegetation-inclusive top surfaces at non-imagery epochs, and the reported
records carry the scope heterogeneity of Sec.~\ref{sec:data_reference}.
Third, the
region-level calibration rests on $n = 10$ regions; its uncertainty is
expressed as
the 90\% prediction interval, spanning a factor of about 3.6 in either
direction, and the low-reliability
tier,
within
which dispersed-debris regimes (Orange Beach; the failure column of
Fig.~\ref{fig:spatial}) remain the hardest cases. Fourth, per-pixel spatial
correlation with the drone surface is modest (median $\approx$0.34). The
method measures integrated cell volume well, not the exact placement of
every pile, and applications should consume it at cell level. Fifth, one
reference (Grand Isle) is biased by cleanup timing, which we disclose and
read directionally, treating its fused reference as a lower bound, rather
than
excluding the region.

\section{Conclusion}\label{sec:conclusion}
This paper showed that spatially explicit hurricane debris
volume can be estimated from a single pass of post-event aerial RGB
imagery, completing, together with our prior debris segmentation work, the
arc from detection to quantification. DebrisHeightNet conditions a
lightweight
trained head on two frozen vision foundation models to regress per-pixel
debris height. CW-LMF synthesizes the height supervision that no
post-hurricane
sensor provides at the imagery epoch. A region-level power law then
calibrates image-derived
volume to reported
hauled debris, the quantity that the organizations responsible
for debris
management and removal, from local and state agencies to national programs
such as \mbox{FEMA} Public Assistance, plan, contract, and reimburse
against. The calibration carries an explicit 90\% prediction interval
spanning a factor of about 3.6 in either direction, and it reduces the
median leave-one-out multiplicative error from
2.7$\times$ to 1.9$\times$.
Evaluated across ten regions, five hurricanes, and three
states, the uncalibrated model lands
within 30\% of the
reported record at the
training region
where the Hazus and FEMA-hybrid forecasts over-predict severalfold, and
agrees with an
independent UAV survey at Spearman $\rho = 0.87$.

Four directions for future research follow directly from the limitations. Per-sensor height
calibration is the natural remedy for the scale drift that deployment on a
new imagery
source can introduce. Evaluation on further imagery sources, including
satellite platforms, would broaden the method's reach, and the
drone cross-check already shows encouraging agreement against an
independently sensed reference. Broader storm regimes,
especially dispersed-debris events, would extend the calibration's support.
And a LiDAR campaign flown together with the first post-storm imagery
would remove the temporal mismatch that currently limits the training
target's corroboration.

Beyond these method extensions, the estimates can feed the predictive and
mitigation practice of the natural-hazards field. Measured per-cell volumes
provide event-scale observations for validating and recalibrating the
parametric debris-forecast models that regional risk and recovery
simulations rely on today. In mitigation and response planning, mapped
debris concentrations can inform the siting of temporary collection areas,
the scoping of pre-positioned removal contracts, and post-event checks of
hauling progress against what the imagery shows on the ground.

\section*{Acknowledgments}
Kooshan Amini and Jamie E. Padgett were partially supported by NSF Award
2227467 and by the Ken Kennedy Institute's inaugural Research Clusters
Initiative. Any opinions, findings, and conclusions or
recommendations expressed in this material are those of the authors and do
not necessarily reflect the views of the National Science Foundation. The
authors thank Lee County, Florida, for the public-records release of
per-truck debris collection data.

\section*{Data Availability}
NOAA Emergency Response Imagery is publicly available from NOAA
(\url{https://storms.ngs.noaa.gov}); post-event LiDAR and pre-event DEMs are
distributed through NOAA Digital Coast and USGS 3DEP; the UAV survey is
published on DesignSafe as project PRJ-4211~\cite{designsafe_prj4211}, and
the building-damage assessments used by the parametric comparison as project
PRJ-5700~\cite{designsafe_prj5700}. The
reported hauled-debris catalogue (Supplementary
Table~\ref{tab:real_debris_sources}), per-region model
outputs, the trained checkpoint, configurations, and code will be made
available upon publication.

\bibliographystyle{IEEEtran}
\bibliography{references}

\clearpage
\onecolumn
\raggedbottom 
\setcounter{figure}{0}
\renewcommand{\thefigure}{S\arabic{figure}}
\renewcommand{\theHfigure}{S\arabic{figure}}
\setcounter{table}{0}
\renewcommand{\thetable}{S\arabic{table}}
\renewcommand{\theHtable}{S\arabic{table}}
\setcounter{equation}{0}
\renewcommand{\theequation}{S\arabic{equation}}
\renewcommand{\theHequation}{S\arabic{equation}}

\begin{center}
{\LARGE\scshape Supplementary Information}\\[0.4em]
{\large Rapid Debris-Volume Estimation
from Post-Hurricane Aerial Imagery}
\end{center}
\vspace{0.8em}

\noindent This supplement provides the complete training-loss specification
(Sec.~A), the CW-LMF configuration and a worked example (Sec.~B,
Fig.~\ref{fig:si_cwlmf}), the fusion redistribution decomposition (Sec.~C,
Fig.~\ref{fig:si_redistribution}), the monocular-backbone comparison
(Sec.~D, Fig.~\ref{fig:si_backbone}), the calibration predictor and
functional-form comparisons (Sec.~E, Fig.~\ref{fig:si_calibration}), and the
per-region provenance and reported-record catalogues (Sec.~F,
Tables~\ref{tab:provenance} and~\ref{tab:real_debris_sources}).

\subsection*{A. Training Loss: Complete Specification}
The loss of Sec.~\ref{sec:method_loss} is specified below; all pixel terms
operate in log space and the volume term in linear space. For each valid
(water-masked) pixel $p$, let $h_p$ be the fused-target height in meters,
$\ell_p = \log\!\left(1 + \max(h_p, 0)\right)$ its log-space target, and
$(\mu_p, \log\sigma^2_p)$ the model output, with
$\sigma^2_p = \exp(\log\sigma^2_p)$ and the log-variance clamped to $[-6,6]$
by the head (Sec.~\ref{sec:method_dhn}). Every pixel carries the combined
class and height-proportional weight of
Sec.~\ref{sec:method_loss}
\begin{equation}
w_p \;=\; c_{s_p}\left(1 + 0.5\,\ell_p\right),
\qquad c = (1,\,8,\,12)
\label{eq:si_weights}
\end{equation}
for CLIPSeg-debris class $s_p \in$ \{no, low-density, high-density debris\},
with $W = \sum_p w_p$ over the valid pixels of a batch. The heteroscedastic
Gaussian negative log-likelihood and the asymmetric MAE
of Sec.~\ref{sec:method_loss} are
\begin{equation}
\mathcal{L}_{\mathrm{NLL}} \;=\; \frac{1}{W}\sum_{p} w_p\,
\tfrac{1}{2}\!\left[\log\sigma^2_p +
\frac{(\ell_p - \mu_p)^2}{\sigma^2_p}\right],
\label{eq:si_nll}
\end{equation}
\begin{equation}
\mathcal{L}_{\mathrm{asym}} \;=\; \frac{1}{W}\sum_{p} w_p\, a_p\,
\left|\ell_p - \mu_p\right|,
\quad
a_p =
\begin{cases}
2.5 & \mu_p < \ell_p,\\
1 & \text{otherwise},
\end{cases}
\label{eq:si_asym}
\end{equation}
so that under-prediction is penalized $2.5\times$ more than over-prediction.
The log-variance regularizer is the unweighted mean over valid
pixels,
\begin{equation}
\mathcal{R}_{\mathrm{var}} \;=\; \frac{1}{|V|}\sum_{p \in V}
\left(\log\sigma^2_p\right)^2 .
\label{eq:si_logvar}
\end{equation}
The gradient-matching term compares forward differences of the
prediction and the log-space target over horizontally and vertically adjacent
valid-pixel pairs,
\begin{equation}
\mathcal{L}_{\mathrm{grad}} \;=\;
\mathrm{mean}_{x}\!\left|\Delta_x\mu - \Delta_x\ell\right| \;+\;
\mathrm{mean}_{y}\!\left|\Delta_y\mu - \Delta_y\ell\right| .
\label{eq:si_grad}
\end{equation}
The differentiable per-cell volume term converts predictions back
to linear height and compares integrated sums over the cell's
debris-classified valid pixels,
\begin{equation}
\mathcal{L}_{\mathrm{vol}} \;=\; \operatorname*{mean}_{\text{cells}}\,
\min\!\left(\frac{\bigl|\hat{V} - \tilde{V}\bigr|}
{\tilde{V} + \varepsilon},\; 5\right),
\label{eq:si_vol}
\end{equation}
with $\hat{V} = \sum_p \max\!\left(e^{\mu_p}-1,\,0\right)$ and
$\tilde{V} = \sum_p \max(h_p, 0)$ in pixel-height units (the shared pixel
footprint cancels in the ratio), $\varepsilon = 10^{-6}$, and the per-cell
clamp at 5 capping outlier gradients. The total loss is
\begin{equation}
\mathcal{L} \;=\; \mathcal{L}_{\mathrm{NLL}}
+ 2.0\,\mathcal{L}_{\mathrm{asym}}
+ 0.05\,\mathcal{R}_{\mathrm{var}}
+ 0.5\,\mathcal{L}_{\mathrm{grad}}
+ 0.02\,\mathcal{L}_{\mathrm{vol}},
\label{eq:si_total}
\end{equation}
with the volume term enabled only after a 30-epoch warm-up so the pixel
losses stabilize predictions before the integrated signal is introduced.

\subsection*{B. CW-LMF Configuration and Worked Example}
\begin{figure}[H]
\centering
\includegraphics[width=\textwidth]{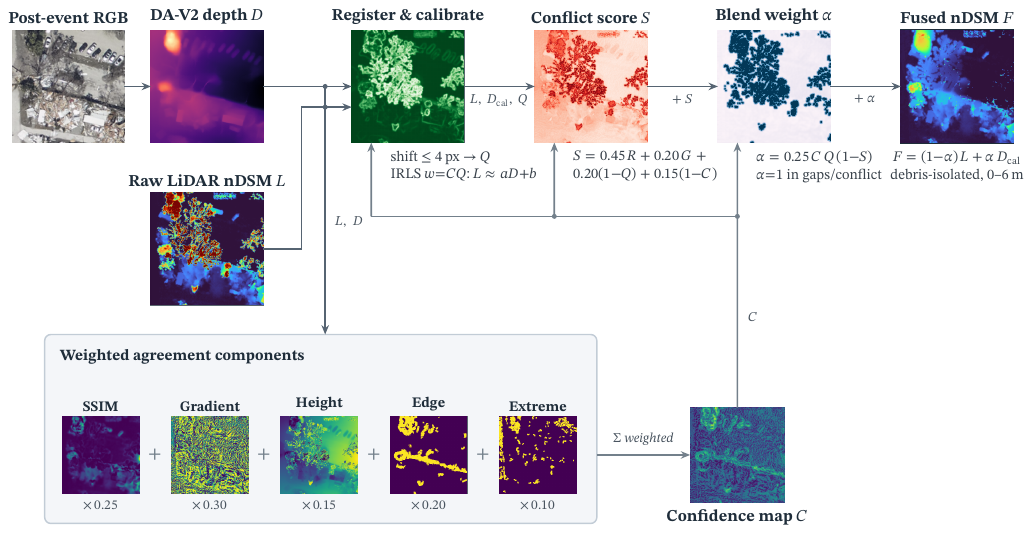}
\caption{CW-LMF worked example on one Estero training cell, with a real map
at every stage of the fusion (Sec.~\ref{sec:method_cwlmf}): inputs $L$ and
$D$, registration and calibration, agreement components and confidence $C$,
conflict score $S$, blend weight $\alpha$, and fused target $F$.}
\label{fig:si_cwlmf}
\end{figure}

Fig.~\ref{fig:si_cwlmf} walks the full CW-LMF computation through one real
training cell: the raw LiDAR nDSM $L$ and the relative depth $D$ enter the
fusion spine (register \& calibrate $\rightarrow$ conflict score $S$
$\rightarrow$ blend weight $\alpha$ $\rightarrow$ fused target $F$;
Eqs.~(\ref{eq:conflict})--(\ref{eq:fuse})) and in parallel feed the five
agreement components whose weighted sum is the per-pixel confidence map $C$,
which in turn enters the calibration weights, the conflict score, and the
blend weight. Beyond the quantities defined in
Sec.~\ref{sec:method_cwlmf}, the deployed configuration fixes: maximum
integer registration shift 4\,px (grid search maximizing gradient
correlation, yielding $Q$); robust-calibration weights
$w = \mathrm{clip}(C\,Q,\,0.02,\,1)$ in the iteratively reweighted fit of
$D_{\mathrm{cal}} = aD + b$; conflict-score weights of
Eq.~(\ref{eq:conflict}) $w_h = 0.45$, $w_g = 0.20$, $w_q = 0.20$,
$w_c = 0.15$; conflict mask from an adaptive threshold on the
conflict score $S$, with floor 0.15
($\max(0.15,\ \mathrm{median} + 1.5\,\mathrm{MAD})$); maximum monocular share
0.25 in agreeing interior overlap [Eq.~(\ref{eq:alpha}); boundary feathering
between regimes can locally exceed it]; blend feathering over a
10\,px transition band with Gaussian smoothing $\sigma = 1.5$\,px of the
$\alpha$ map; and a minimum valid fraction of 0.05 for a cell to be
processed. The per-pixel confidence $C$ is the weighted sum of five agreement
components (structural similarity 0.25, gradient correlation 0.30, height
agreement 0.15, edge overlap 0.20, stable-extreme bonus 0.10), where height
agreement is $1 - \mathrm{clip}(2\,|\Delta h|)$ on normalized heights. As
stated in Sec.~\ref{sec:method_cwlmf}, this configuration was fixed
from the design principles and fusion diagnostics of the training region,
before any evaluation-region comparison, and
never tuned against any evaluation quantity.

\subsection*{C. Fusion Redistribution by Class and Height}
\begin{figure}[H]
\centering
\includegraphics[width=\textwidth]{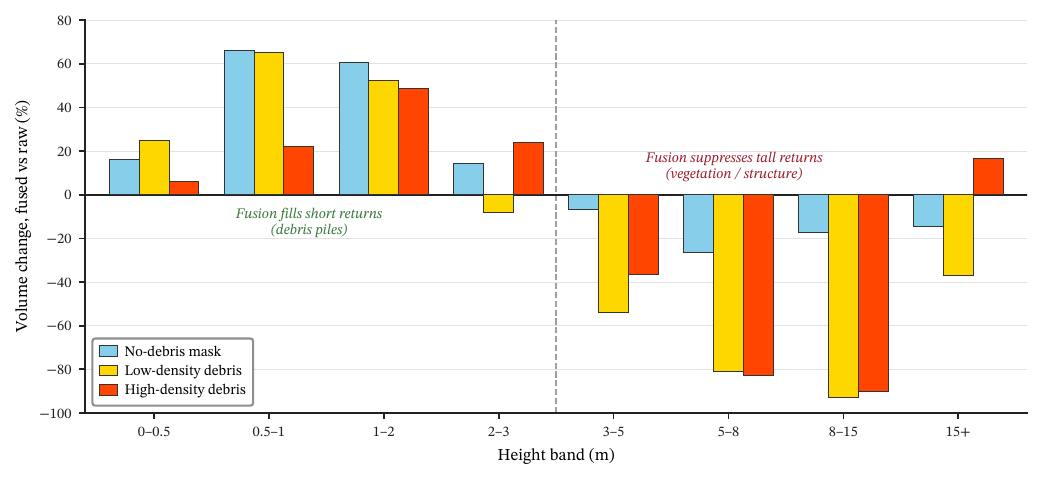}
\caption{CW-LMF volume redistribution by CLIPSeg-debris class and height
band: signed change in class volume, (fused $-$ raw)/raw, per height band,
over 3{,}490 cells of the Estero evaluation extent.}
\label{fig:si_redistribution}
\end{figure}

Fig.~\ref{fig:si_redistribution} is the decomposition cited in the mechanism
discussion (Sec.~\ref{sec:discussion}): the signed change in class volume
between the fused target and the raw LiDAR nDSM, per height band and
CLIPSeg-debris class, over 3{,}490 cells of the Estero evaluation extent
(the cells of the Estero grid block with valid raw and fused surfaces,
between the 3{,}178-cell ROI of Table~\ref{tab:master} and the 3{,}645
split cells of Sec.~\ref{sec:method_training}).
A cell contributes to every class it contains
(3{,}350 / 2{,}175 / 1{,}116 cells for no-/low-/high-density debris). The
headline behavior, raising the short debris bands and suppressing the
tall vegetation and structure bands, is the height-selective redistribution
the design intends rather than a uniform smoothing of the LiDAR; the per-band
numbers quoted in Sec.~\ref{sec:discussion} come from this decomposition.
The net per-class fused-to-raw volume ratios, a different statistic from the
per-band bars, are 0.91 (no-debris), 0.65 (low-density), and 0.86
(high-density). The isolated 15+\,m high-density gain is a
small-sample effect ($\approx$7{,}700\,px, the sparsest tall band) where
monocular depth fills the apex of the tallest dense piles.

\subsection*{D. Monocular-Backbone Comparison}
\begin{figure}[H]
\centering
\includegraphics[width=\textwidth]{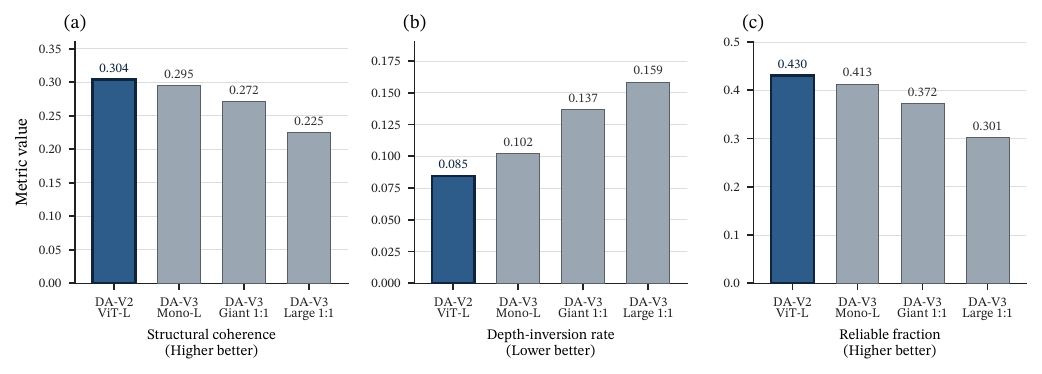}
\caption{Monocular-backbone comparison on NOAA Estero tiles
($n = 3{,}854$): (a) structural coherence (higher is better), (b)
depth-inversion rate (lower is better), and (c) reliable fraction (higher is
better; all three defined in the text) for DA-V2 ViT-L against three Depth
Anything~3 variants.}
\label{fig:si_backbone}
\end{figure}

Fig.~\ref{fig:si_backbone} reports the backbone assessment behind the
DA-V2-over-successor choice (Secs.~\ref{sec:method_dhn}
and~\ref{sec:results_ablations}), computed on $n = 3{,}854$ NOAA Estero
tiles with the raw LiDAR nDSM as the reference surface. Three
reference-based metrics summarize each candidate's suitability.
\emph{Structural coherence} thresholds both the LiDAR nDSM and the depth map
into elevated- and low-relief masks and scores their weighted
intersection-over-union, measuring how faithfully the depth map reproduces
the LiDAR's structural footprints. The \emph{depth-inversion rate} is the
fraction of tiles whose depth map correlates negatively with the LiDAR
surface, i.e., whose near--far sense is reversed. The \emph{reliable
fraction} is the mean share of pixels per tile whose LiDAR--depth agreement
confidence (the five-component confidence of Sec.~B) exceeds a fixed
threshold.
DA-V2 ViT-L attains the highest structural coherence (0.304 versus
0.295 for the best Depth Anything 3 variant), the lowest depth-inversion
rate (0.085 versus 0.102), and the highest reliable fraction (0.430 versus
0.413), and is the adopted backbone.

\subsection*{E. Calibration Predictor and Functional-Form Comparisons}
\begin{figure}[H]
\centering
\includegraphics[width=\textwidth]{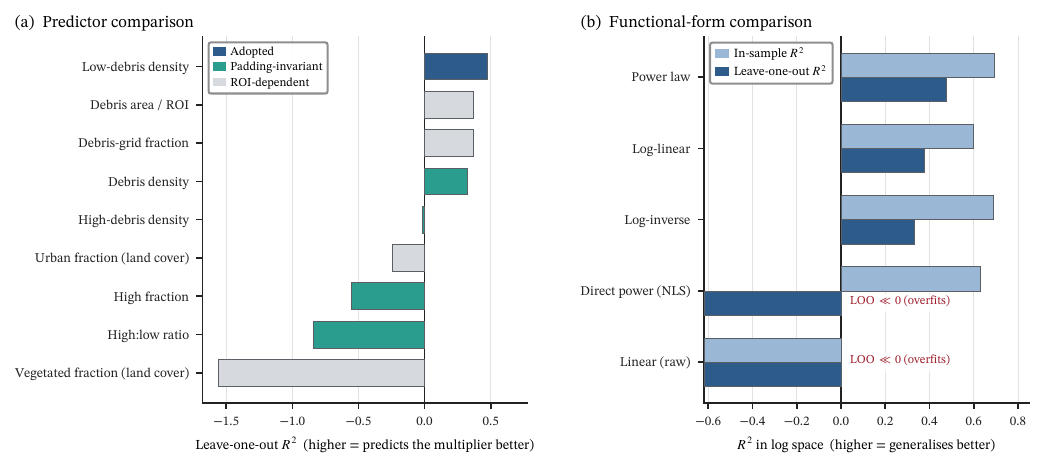}
\caption{Calibration design choices ($n = 10$ regions; leave-one-out $R^2$
as the criterion): (a) predictor comparison; (b) functional-form
comparison.}
\label{fig:si_calibration}
\end{figure}

Fig.~\ref{fig:si_calibration} details the two selection studies summarized
in Sec.~\ref{sec:method_calibration}, both on the $n = 10$ region set with
leave-one-out (LOO) $R^2$ as the criterion. Among candidate predictors of
the reported-to-model multiplier, the adopted low-density
debris fraction $m$ is invariant to debris-free ROI padding and attains the
best LOO $R^2$
($+0.48$); area-normalized fractions are ROI-dependent and weaker; and two
Esri land-cover covariates, fitted through the same log--log power-law and
LOO machinery, lose clearly (vegetated LOO $-1.56$; urban LOO $-0.25$).
Predictors that share the model volume term with the target ratio (mean pile
height, raw volume per debris cell) are excluded a priori. Among functional
forms, the log--log power law generalizes best, while raw-space fits overfit
(LOO collapses below zero) and serve as negative controls. Because the
predictor and functional form are selected on the same ten regions used for
the leave-one-out evaluation, the reported LOO $R^2$ is selection-optimistic;
no external calibration holdout exists at this sample size, and region-wise
leave-one-out leaves same-hurricane and same-source neighbors in the
training folds.

\subsection*{F. Region Provenance and Reported-Record Catalogue}
Table~\ref{tab:provenance} records, for each region, the NOAA ERI flight
(bucket date and days after landfall), the post-event LiDAR campaign
(provider, NOAA Digital Coast identifier, and acquisition-window lag), the
pre-event DEM, the coordinate reference systems, and the realized
fused-LiDAR coverage of the debris cells.
Table~\ref{tab:real_debris_sources} catalogues the reported hauled-debris
records behind Table~\ref{tab:master}: the reconciled median used throughout
the paper, the min--max span across all catalogued sources, the number of
contributing sources, and the scope, vegetation share, category, and
citation of each region's primary record. The tables' notes carry the
provenance caveats that matter for interpretation, chief among them Grand
Isle's 13-month-plus LiDAR lag and Treasure Island's combined
Helene--Milton record.

\begin{table}[H]
\centering
\caption{Region and data provenance: NOAA Emergency Response Imagery (ERI)
flight, post-event LiDAR campaign, pre-event DEM, coordinate reference
systems (CRS; EPSG codes), and realized fused-LiDAR coverage of the debris
cells.}
\label{tab:provenance}
\footnotesize
\setlength{\tabcolsep}{5pt}
\renewcommand{\arraystretch}{1.15}
\begin{threeparttable}
\begin{tabular}{@{}l l l c l c r@{}}
\toprule
\textbf{Region} & \textbf{NOAA ERI flight} &
\textbf{Post-event LiDAR} & \textbf{LiDAR} & \textbf{Pre-event DEM} &
\textbf{Target} & \textbf{Cov.} \\
 & \textbf{(days post)} & \textbf{(NOAA DC ID; lag)} & \textbf{CRS} &
\textbf{(ID, year, res.)} & \textbf{CRS} & \textbf{\%} \\
\midrule
\multicolumn{7}{@{}l}{\emph{Hurricane Ian (2022) -- Florida}\tnote{a}} \\
Pine Island    & 20220930d (2\,d) & USACE; DC\,9651 (11--24\,d) & 6346 & USGS FL SW (9049, 2018, 1\,m) & 26917 & 0.0 \\
Estero Island  & 20220930d (2\,d) & USACE; DC\,9651 (11--24\,d) & 6346 & USGS FL SW (9049, 2018, 1\,m) & 26917 & 100.0 \\
San Carlos Is. & 20220930d (2\,d) & USACE; DC\,9651 (11--24\,d) & 6346 & USGS FL SW (9049, 2018, 1\,m) & 26917 & 99.8 \\
Captiva Island & 20220930a (2\,d) & USACE; DC\,9651 (11--24\,d) & 6346 & USGS FL SW (9049, 2018, 1\,m) & 26917 & 100.0 \\
Iona Nbhd.     & 20220930d (2\,d) & USACE; DC\,9651 (11--24\,d) & 6346 & USGS FL SW (9049, 2018, 1\,m) & 26917 & 5.6 \\
\midrule
\multicolumn{7}{@{}l}{\emph{Hurricane Michael (2018) -- Florida}} \\
Panama City    & 20181011a (1\,d) & USACE; DC\,8625 (14--25\,d) & 6318 & NOAA Lower Choctaw.\ (8682, 2017, 1\,m)\tnote{b} & 26916 & 0.0 \\
Mexico Beach   & 20181011a (1\,d) & USACE; DC\,8625 (14--25\,d) & 6318 & NOAA Lower Choctaw.\ (8682, 2017, 1\,m)\tnote{b} & 26916 & 91.7 \\
\midrule
\multicolumn{7}{@{}l}{\emph{Hurricane Ida (2021) -- Louisiana}} \\
Grand Isle\tnote{c} & 20210830a (1\,d) & NGS; DC\,10200 (401--418\,d) & 6344 & USGS LA Coastal (3DEP, 2020, 1\,m) & 26915 & 100.0 \\
\midrule
\multicolumn{7}{@{}l}{\emph{Hurricane Sally (2020) -- Alabama}} \\
Orange Beach   & 20200918a (2\,d) & USACE; DC\,9200 (9--27\,d)\tnote{d} & 6345 & NGS NW FL Topobathy (9708, 2020, 1\,m) & 26916 & 23.1 \\
\midrule
\multicolumn{7}{@{}l}{\emph{Hurricane Milton (2024) -- Florida}} \\
Treasure Island & 20241011d (2\,d) & USACE; DC\,10196 (16--22\,d)\tnote{d} & 6346 & USGS FL Peninsular (3DEP, 2018, 1\,m) & 26917 & 66.6 \\
\bottomrule
\end{tabular}
\begin{tablenotes}[flushleft]\scriptsize
\item NOAA ERI flight is the bucket date (\texttt{YYYYMMDD\{a..d\}}) of the
earliest available post-event flight over each region. Landfalls: Ian
2022-09-28; Michael 2018-10-10; Ida 2021-08-29; Sally 2020-09-16; Milton
2024-10-09. LiDAR ``lag'' is days from landfall spanning the campaign's
documented acquisition window; all pre-event
DEMs are 1\,m bare-earth (NAVD88).
\item[a] Ian lags span the per-tile acquisition dates carried in the
delivered DC\,9651 files (2022-10-09 to 10-22 across the campaign; per-point
GPS times over Estero Island span Oct.\ 11--20, with 99\% of returns on
Oct.\ 11--13). The campaign's published temporal extent (2022-11-16 to
11-18) does
not match the delivered tiles, so the file-level dates are reported.
\item[b] The 2017 Lower Choctawhatchee DEM ships elevations in US survey
feet (converted to meters).
\item[c] Grand Isle's post-Ida LiDAR was flown more than 13 months
post-landfall, after partial debris removal, so its fused reference
under-captures the pre-removal debris.
\item[d] Post-Sally and post-Milton windows are the collection periods
from official NOAA metadata for the blocks covering each study area (Sally:
the Alabama block; Milton: the west-coast blocks); both are coastal-only
campaigns, explaining the partial
coverage.
\end{tablenotes}
\end{threeparttable}
\end{table}

\begin{table}[H]
\centering
\caption{Reported-debris reference sources. Reported (median): the
per-region value used in Table~\ref{tab:master}. Source span: min--max
across all catalogued sources. $n$: number of reconciled estimates. Scope,
Veg.\,\%, source category, and citation describe each region's primary
record.}
\label{tab:real_debris_sources}
\footnotesize
\setlength{\tabcolsep}{5pt}
\renewcommand{\arraystretch}{1.15}
\begin{threeparttable}
\begin{tabular}{@{}l r c c c r l l@{}}
\toprule
\textbf{Region} & \textbf{Reported} & \textbf{Source span} & \textbf{n} &
\textbf{Scope} & \textbf{Veg.} & \textbf{Source category} & \textbf{Citation} \\
 & \textbf{(median, m\textsuperscript{3})} & \textbf{(m\textsuperscript{3})} &
 & & \textbf{\%} & \textbf{(primary record)} & \\
\midrule
\multicolumn{8}{@{}l}{\emph{Hurricane Ian (2022)}} \\
Pine Island            & 442{,}576 & 349{,}962--535{,}189     & 2 & roi  & 58.1 & Lee Co.\ public records (per-truck)      & Lee Cty.\ records \\
Estero Island\tnote{a} & 187{,}868 & 75{,}995--4{,}408{,}553  & 3 & roi  & 3.5  & Lee Co.\ public records (per-truck)      & Lee Cty.\ records \\
San Carlos Is.         & 100{,}122 & ---                      & 1 & roi  & 0.7  & Lee Co.\ public records (per-truck)      & Lee Cty.\ records \\
Captiva Island         & 52{,}186  & ---                      & 1 & roi  & 69.0 & Lee Co.\ public records (per-truck)      & Lee Cty.\ records \\
Iona Nbhd.             & 342{,}667 & ---                      & 1 & roi  & 40.7 & Lee Co.\ public records (per-truck)      & Lee Cty.\ records \\
\midrule
\multicolumn{8}{@{}l}{\emph{Hurricane Michael (2018)}} \\
Mexico Beach\tnote{b}  & 373{,}134 & 225{,}768--605{,}968     & 3 & roi  & 0.0  & FEMA FIDA App~9343 (C\&D)      & FIDA (Dec~2022) \\
Panama City            & 857{,}705 & 844{,}934--4{,}357{,}963 & 3 & city & 51.9 & FEMA Public-Assistance release & FEMA, 18~Mar~2021 \\
\midrule
\multicolumn{8}{@{}l}{\emph{Hurricane Ida (2021)}} \\
Grand Isle             & 276{,}587 & 90{,}867--324{,}627      & 3 & roi  & 0.0  & DRC contractor C\&D snapshot   & NOLA.com, 10~Nov~2021 \\
\midrule
\multicolumn{8}{@{}l}{\emph{Hurricane Sally (2020)}} \\
Orange Beach           & 243{,}283 & 110{,}857--374{,}734     & 4 & city & 57.9 & FEMA Public-Assistance release & FEMA, 18~May~2021 \\
\midrule
\multicolumn{8}{@{}l}{\emph{Hurricanes Helene $+$ Milton (2024)}} \\
Treasure Island\tnote{c} & 146{,}795 & 97{,}863--195{,}727    & 2 & city & n/a  & City/news (combined contract)  & Wash.\ Post, 17~Apr~2025 \\
\bottomrule
\end{tabular}
\begin{tablenotes}[flushleft]\scriptsize
\item Scope: roi $=$ spatially filtered to the model region of interest;
city $=$ municipality total. \textbf{Veg.\,\%} is the vegetation share of
the primary haul record; a construction-and-demolition-only record is
already vegetation-free. The median pools sources of mixed scope and
vegetation treatment, so it is not a strictly non-vegetative quantity.
Cubic-yard records are converted to cubic meters. The complete per-source
catalogue is part of the data release (see Data Availability).
\item[a] Estero's span maximum is a broader county-scale figure excluded
from the median; the median (187{,}868) is over the 3 ROI+city estimates, of
which the strict non-vegetative ROI value is 181{,}224.
\item[b] Mexico Beach's primary record is a single construction-and-demolition
(C\&D) line item and is
itself vegetation-free; the median also draws on city-scope totals, and the
full-city haul is $\sim$11\,\% vegetation.
\item[c] Treasure Island is a combined Helene$+$Milton emergency contract
with no published vegetation split; the span runs from a news-reported
contract volume to a tonnage-derived estimate.
\end{tablenotes}
\end{threeparttable}
\end{table}

\end{document}